\documentclass[lettersize,10pt,journal,twoside]{IEEEtran}
\usepackage{ieee_robotics}
\usepackage{orcidlink}
\usepackage{physics}

\hypersetup{
  colorlinks=true,
  urlcolor=blue
}
\definecolor{rev1Color}{rgb}{0,0,0}
\definecolor{rev2Color}{rgb}{1,0,0}

\acrodef{qp}[QP]{Quadratic Programming}
\acrodef{fd}[FD]{Force Decomposition}
\acrodef{dof}[DoF]{Degree of Freedom}
\acrodef{ros}[ROS]{Robot Operating System}
\acrodef{uav}[UAV]{Unmanned Aerial Vehicle}
\acrodef{dof}[DoF]{Degree-of-freedom}
\acrodef{com}[CoM]{center of mass}
\acrodef{mrr}[MRR]{Modular reconfigurable robot}
\acrodef{rl}[RL]{Reinforcement Learning}
\acrodef{icp}[ICP]{Iterative Closest Point}

\newcommand{\method}{Tac-Man\xspace}
\newcommand{\transpose}{\mathsf{T}}

\newcommand{\suppUrlSO}{https://vimeo.com/907222643/9793e630b0}
\newcommand{\suppUrlSI}{https://vimeo.com/907224197/f38b1dea62}
\newcommand{\suppUrlSII}{https://vimeo.com/907224440/b059314753}
\newcommand{\suppUrlSIII}{https://vimeo.com/907224671/cb9aa51fcb}
\newcommand{\suppUrlSIV}{https://vimeo.com/907224990/3b4181a4e0}
\newcommand{\suppUrlSV}{https://vimeo.com/907225198/b641a5a84d}

\title{\method: Tactile-Informed Prior-Free\\Manipulation of Articulated Objects}

\author{Zihang Zhao\orcidlink{0000-0003-3215-7152}, Yuyang Li\orcidlink{0000-0002-5794-7997}, Wanlin Li\orcidlink{0000-0002-8538-8571}, Zhenghao Qi\orcidlink{0009-0004-2924-6714}, Lecheng Ruan\orcidlink{0000-0001-5061-3575},\\ Yixin Zhu\orcidlink{0000-0001-7024-1545}, and Kaspar Althoefer\orcidlink{0000-0002-1141-9996}, \textit{Senior Member, IEEE}%

\thanks{This work is supported in part by the National Science and Technology Major Project (2022ZD0114900), the National Natural Science Foundation of China (62376009), the Beijing Nova Program, the State Key Lab of General AI at Peking University, the PKU-BingJi Joint Laboratory for Artificial Intelligence, the National Comprehensive Experimental Base for Governance of Intelligent Society, Wuhan East Lake High-Tech Development Zone, and the Horizon Europe Framework through the PALPABLE project (101092518). (Zihang Zhao and Yuyang Li contributed equally to this work. Corresponding authors: Lecheng Ruan and Yixin Zhu.)
}%
\thanks{Zihang Zhao and Yixin Zhu are with the Institute for Artificial Intelligence, Peking University, Beijing 100871, China (email: \texttt{zhaozihang@stu.pku.edu.cn}; \texttt{yixin.zhu@pku.edu.cn}).}%
\thanks{Yuyang Li and Zhenghao Qi are with the Institute for Artificial Intelligence, Peking University, Beijing 100871, China, and the Beijing Institute for General Artificial Intelligence, Beijing 100080, China, and also with the Department of Automation, Tsinghua University, Beijing 100084, China (emails: \texttt{\{liyuyang20,qi-zh21\}@mails.tsinghua.edu.cn}).}%
\thanks{Wanlin Li is with the Beijing Institute for General Artificial Intelligence, Beijing 100080, China (emails: \texttt{liwanlin@bigai.ai}).}%
\thanks{Lecheng Ruan is with the College of Engineering, Peking University, Beijing 100871, China, and also with the PKU-Wuhan Institute for Artificial Intelligence, Wuhan 430075, China (emails: \texttt{ruanlecheng@ucla.edu}).}%
\thanks{Zihang Zhao and Lecheng Ruan were also with the Beijing Institute for General Artificial Intelligence partially during this work.}
\thanks{Kaspar Althoefer is with the Centre for Advanced Robotics @ Queen Mary, within the School of Engineering \& Materials Science, Queen Mary University of London, London E1 4NS, UK (email: \texttt{k.althoefer@qmul.ac.uk}).}%
\thanks{Digital Object Identifier (DOI): 10.1109/TRO.2024.3508134.}%
}

\begin{document}

\let\oldtwocolumn\twocolumn
\renewcommand\twocolumn[1][]{%
    \oldtwocolumn[{#1}{
        \vspace{-24pt}
        \centering
        \includegraphics[width=\linewidth]{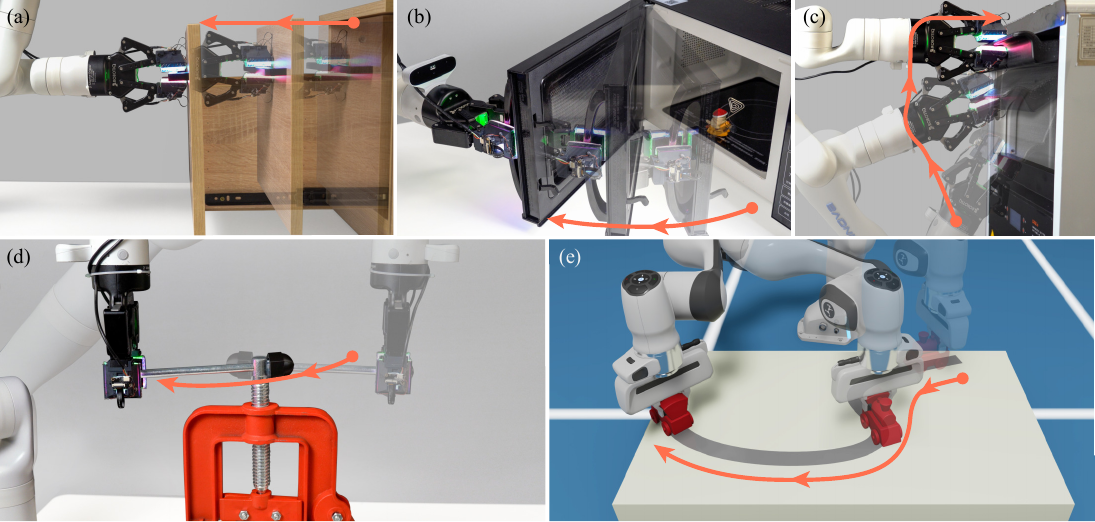}
        \captionof{figure}{\textbf{Tactile-informed prior-free manipulation of articulated objects.}
        Timely compensation for contact deviations allows robots to precisely manipulate a variety of articulated objects, eliminating the need for prior kinematic model knowledge. The capabilities are showcased by (a) manipulating objects with prismatic joints, such as drawers; (b) handling objects with revolute joints, such as microwave ovens; and managing complex mechanisms that involve simultaneous translational and rotational movements, as exemplified by (c) an ice machine and (d) a vise. (e) Additionally, the robot proficiently performs manipulation following arbitrary trajectories in simulation; see \href{\suppUrlSO}{Supp Video S0}. 
        }
        \label{fig:teaser}
    }]
}

\markboth{IEEE Transactions on Robotics.  PREPRINT VERSION. ACCEPTED
November, 2024 }{Zhao \MakeLowercase{\etal}: Tactile-Informed Prior-Free Manipulation of Articulated Objects}

\maketitle

\begin{abstract}
Integrating robots into human-centric environments such as homes, necessitates advanced manipulation skills as robotic devices will need to engage with articulated objects like doors and drawers.
Key challenges in robotic manipulation of articulated objects are the unpredictability and diversity of these objects' internal structures, which render models based on object kinematics priors, both explicit and implicit, inadequate. Their reliability is significantly diminished by pre-interaction ambiguities, imperfect structural parameters, encounters with unknown objects, and unforeseen disturbances.
Here, we present a \textbf{prior-free} strategy, \method, focusing on maintaining stable robot-object contact during manipulation.
Without relying on object priors, \method leverages tactile feedback to enable robots to proficiently handle a variety of articulated objects, including those with complex joints, even when influenced by unexpected disturbances. Demonstrated in both real-world experiments and extensive simulations, it consistently achieves near-perfect success in dynamic and varied settings, outperforming existing methods.
Our results indicate that tactile sensing alone suffices for managing diverse articulated objects, offering greater robustness and generalization than prior-based approaches.
This underscores the importance of detailed contact modeling in complex manipulation tasks, especially with articulated objects.
Advancements in tactile-informed approaches significantly expand the scope of robotic applications in human-centric environments, particularly where accurate models are difficult to obtain. See additional material at \texttt{\url{https://tacman-aom.github.io}}.
\end{abstract}
\begin{IEEEkeywords}
Articulated object manipulation, prior-free, tactile sensing, contact regulation.
\end{IEEEkeywords}

\section{Introduction}\label{sec:intro}

\IEEEPARstart{A}{rticulated} objects, defined by interconnected parts that move in conjunction with each other, include a vast array of everyday items, such as doors, drawers, and appliances~\cite{wang2023rearrange}, as illustrated in \cref{fig:teaser}.
A robot's capability to manipulate articulated objects is essential for effective integration into human-centric environments, assisting humans in diverse tasks while representing a significant challenge within robotics~\cite{krotkov2017darpa,billard2019trends}.

To interact efficiently with these objects, robots traditionally rely on prior knowledge of their kinematics. This knowledge is often embedded either explicitly, as in kinematic models used for designing manipulation strategies via control or planning techniques~\cite{chitta2010planning,burget2013whole,jiao2021consolidating, karayiannidis2012open,hausman2015active,karayiannidis2016adaptive,hu2017learning,abbatematteo2019learning,li2020category,moses2020visual,zeng2021visual,mittal2022articulated,han2022scene,lv2022sagci,jiang2022ditto,eisner2022flowbot3d,martin2022coupled,zhang2023part}, or implicitly, within policy models developed through machine learning techniques~\cite{pastor2009learning,welschehold2017learning,zhang2018deep,lynch2020learning, xiong2021learning,qin2022dexmv,wong2022error,gong2023arnold,ye2023learning,urakami2019doorgym,xu2022universal,chen2022towards,geng2023partmanip,mo2019partnet,xiang2020sapien,liu2022akb,geng2023gapartnet}.
However, acquiring precise prior knowledge about such objects presents 
several spatial and temporal challenges:
\begin{itemize}[leftmargin=*,noitemsep,nolistsep]
    \item \textbf{Ambiguity:} Objects with similar external appearances may have vastly different internal structures, complicating the unique determination of their kinematics through perception. This issue is extensively explored by Zhu \etal~\cite{zhu2020dark}.
    \item \textbf{Imperfectness:} Accurate acquisition of an object's structural parameters is challenging due to perception noise or deviation from standard models, such as the off-axis doors.
    \item \textbf{Unknown:} Certain objects feature unique and sophisticated kinematics, making them difficult to model and generalize, as exemplified by the objects in \cref{fig:teaser}(c) and (d).
    \item \textbf{Obsolescence:} Kinematic models can become obsolescent due to active or passive changes over time, leading to a gap in knowledge when executing manipulation tasks.
\end{itemize}
These challenges prompt us to question the hitherto assumed indispensability of prior knowledge in the manipulation of articulated objects and explore alternative approaches.

Drawing inspiration from human tactile interaction, we propose a novel tactile-informed prior-free manipulation approach, \method. \method is centered around stable robot-object contact during manipulation. By accurately sensing contact geometry information, \method leverages deviations from the initial state to inform the robot of both the direction and magnitude of adjustments needed to effectively manipulate the articulated object. Without any prior-based modeling, \method addresses the challenges posed by ambiguity, imperfectness, unknowns, and obsolescence in object kinematics, which are prevalent in dynamic human-centric environments.

Our comprehensive validation includes a series of real-world experiments as well as extensive simulation studies. The experiments demonstrate \method's proficiency in handling objects with different joint types, from basic prismatic and revolute joints to complex mechanisms with simultaneous translational and rotational movements. We conducted further simulations to scale up these real-world experiments and probe \method's adaptability and robustness. Collectively, these studies confirm the effectiveness of \method, especially in scenarios where traditional reliance on priors is unreliable or inadequate.

The present paper makes three major contributions:
\begin{itemize}[leftmargin=*,noitemsep,nolistsep]
    \item We introduce \method, a novel tactile-informed prior-free manipulation framework for articulated objects. Unlike traditional methods that depend on prior knowledge of object kinematics, \method leverages tactile feedback to dynamically maintain stable contact during the manipulation process. This strategy ensures robust interaction with objects by adapting to real-time tactile data, offering a novel perspective on robotic manipulation.
    \item Through comprehensive real-world experiments, we validate the superior performance of \method over conventional methods in scenarios characterized by ambiguous, imperfect, unknown, and obsolescent priors. These experiments not only underscore \method's adaptability and precision in handling complex manipulations but also demonstrate its potential in enhancing robotic autonomy in unpredictable environments.
    \item We conduct an extensive simulation study to evaluate the scalability and generalization capabilities of \method across a diverse range of articulated objects and trajectories, including those arbitrarily generated. This study, one of the largest of its kind, confirms \method's effectiveness in a broad spectrum of settings, illustrating its versatility and the feasibility of its application in real-world scenarios beyond the confines of controlled experiments.
\end{itemize}

The paper is organized to provide a thorough exploration of \method and its implications. \cref{sec:related_work} reviews relevant literature, positioning our work within the broader context of robotic manipulation of articulated objects. \cref{sec:methods} details \method, our novel tactile-informed, prior-free methodology. \cref{sec:real-world_exp} describes the setup and results of our real-world experiments, while \cref{sec:simulation} presents our simulation studies, showcasing the scalability and generalizability of \method. In \cref{sec:discussion}, we present a detailed discussion on the strengths, limitations, and potential applications of \method, as well as considerations for future research. Finally, \cref{sec:conclusion} concludes the paper, summarizing our key findings and contributions to this field.

\section{Related Work}\label{sec:related_work}

\subsection{Manipulation with Explicit Priors}\label{sec:rw:explicit-prior}

Traditional robotic manipulation strategies have predominantly utilized manually predefined kinematic models, supplied externally, for interaction with articulated objects~\cite{chitta2010planning,burget2013whole,jiao2021consolidating}. While this approach has been effective in certain controlled scenarios, it inherently limits the autonomy of the robotic system due to its reliance on pre-established, accurate kinematic information.

Seeking to enhance system autonomy, recent developments in robotic manipulation have shifted focus toward autonomously deriving object models via robot perception. This trend is exemplified by strategies employing visual inputs to predict kinematic models~\cite{hu2017learning,abbatematteo2019learning,li2020category,zeng2021visual,mittal2022articulated,han2022scene,ma2023sim2real}. Nonetheless, this reliance on visual perception introduces ambiguity, as objects with similar appearances can harbor different internal articulations, challenging the accuracy of model predictions~\cite{zhu2020dark,zeng2021visual}. To address these issues, several studies have integrated multi-frame analysis and wrist-mounted force/torque sensors for more accurate kinematic model estimation~\cite{hausman2015active,moses2020visual,martin2022coupled,eisner2022flowbot3d,lv2022sagci,jiang2022ditto,karayiannidis2012open,karayiannidis2016adaptive}. Despite these improvements, the dependency on being able to identify specific joint types, such as revolute or prismatic, limits the efficacy of such methods.

Our work introduces a novel prior-free approach to address these limitations inherent in manipulation methods with explicit priors. \method, validated by both real-world experiments and simulations, demonstrates superior adaptability and efficacy, especially in scenarios where traditional methods, reliant on explicit priors, are found to be unreliable or insufficient.

\subsection{Manipulation with Implicit Priors}\label{sec:rw:implicit-prior}

Robotic manipulation utilizing implicit priors typically relies on extensive data to infer appropriate actions from the observed state of an object. There are two prominent approaches: imitation learning and reinforcement learning.

Imitation learning enables robots to acquire implicit priors through the observation of human demonstrations~\cite{pastor2009learning,welschehold2017learning,zhang2018deep,lynch2020learning, xiong2021learning,qin2022dexmv,wong2022error,gong2023arnold,ye2023learning}, sometimes facilitated by tele-operational systems~\cite{qin2023anyteleop}. One major challenge in this approach is the collection of diverse and rich data from humans~\cite{zheng2022imitation}. Although transfer methods~\cite{qin2022one,tekden2023grasp,li2023gendexgrasp} have improved the applicability to similar objects, extending these priors to a wider range of objects remains a substantial hurdle.

In contrast, reinforcement learning allows robots to develop implicit priors through iterative interactions in simulated settings~\cite{urakami2019doorgym,xu2022universal,chen2022towards,geng2023partmanip}. The advent of extensive datasets of articulated objects~\cite{mo2019partnet,xiang2020sapien,liu2022akb,geng2023gapartnet} has significantly propelled this approach forward. However, the variance in these datasets often falls short in terms of replicating the complexity encountered in real-world situations, leading to incomplete prior acquisition, in turn constraining the robots' manipulation capabilities.

While both approaches show promise in articulated object manipulation, they grapple with the issue of incomplete priors due to limited data coverage. \method, sidestepping the need for laborious data collection and extensive training, addresses these challenges and has demonstrated its efficacy in manipulating a diverse array of articulated objects.

\subsection{Robot-Object Contact Modeling}

Current research tends to oversimplify and overlook the contact between robots and objects. The assumption of a fixed contact point is prevalent in numerous models~\cite{chitta2010planning,burget2013whole,mittal2022articulated,hausman2015active,moses2020visual,lv2022sagci,jiang2022ditto,eisner2022flowbot3d,martin2022coupled}, yet this is often not feasible in practical applications. \cite{jiao2021consolidating} have introduced the concept of a virtual joint~\cite{likar2014virtual} to represent this contact, but in actual implementation, they still rely on the assumption of a fixed joint. Works by~\cite{karayiannidis2012open,karayiannidis2016adaptive} have relaxed this constraint to some extent, permitting only specific types of movement between the robot and the object. \cite{schmid2008opening} estimates contact using tactile sensors. However, it is limited to a specific type of handling based on rational motion and overlooks the potential for slippage.

Our research, however, uncovers that the data obtained from the contact site itself is adequate to support the task of articulated object manipulation, enabling a prior-free approach. By utilizing advanced tactile sensors to regulate contact, \method successfully demonstrates the ability to manipulate articulated objects under a diverse array of scenarios. This prior-free approach significantly enhances robots' manipulation skills in comparison to existing methodologies.

\subsection{Contact Sensing}

Sensing of robot-object contacts is crucial for contact regulation~\cite{dahiya2009tactile,dahiya2013robotic,dahiya2013directions,yousef2011tactile,kaboli2018robust,lloyd2023pose}. Decades of development have yielded various sensing technologies regarding this issue, broadly categorized as intrinsic or extrinsic styles. Intrinsic tactile sensors, such as 6D force/torque sensors~\cite{ATI}, are integrated within the robot's structure, providing indirect contact measurements through internal force sensing. Extrinsic tactile sensors, attached to the robot's interacting surface, often measure contact information in a more direct manner, including conductive-fluid-based (\eg, BioTac~\cite{fishel2012sensing}), resistor-based~\cite{yang2021non,yu2022all}, piezoelectricity-based~\cite{tian2019rich,lv2022flexible}, capacitance-based~\cite{boutry2018hierarchically,dawood2023learning}, and vision-based~\cite{yuan2017gelsight,ward2018tactip,li2020f,li20233} sensors, each with unique principles for measuring robot-object contacts.

The effectiveness of \method relies on timely compensation for contact deviations, which are modeled geometrically as the changes in the contact point positions, effectively informing both the direction and the magnitude of the necessary adjustments. Under this formulation, sensors with a primary focus on sensing contact geometry, such as GelSight~\cite{yuan2017gelsight}, TacTip~\cite{ward2018tactip}, and others, can be directly adopted within the \method framework. Additionally, sensors primarily designed to measure contact forces can also be utilized within our \method framework, provided that they are properly calibrated according to Hooke's law~\cite{gould1994introduction} to obtain geometric information. A detailed discussion on sensor selection is available in \cref{sec:discussion}.

In this research, we adopt the GelSight-type sensor for the pipeline illustration and experiment verification of the \method among the set of sensors with geometric measurements as one economical option. However, it should be noted that there are no methodological preferences among these sensors with geometric measurement ability, and the formulations should be very similar to each other, with only slight differences in the pre-processing procedure where contact deviations are extracted from the raw sensor data.

\subsection{Tactile Servoing}\label{sec:tactile_servoing}

Tactile servoing, which involves adjusting robot poses in response to tactile signals, has been a significant area of research since its inception by Weiss \etal~\cite{weiss1987dynamic} and first implementation by Berger and Khosla~\cite{berger1991using}. It has evolved into two main approaches: model-based and learning-based. Model-based tactile servoing uses explicit models such as the inverse tactile model~\cite{chen1995edge}, the tactile Jacobian model~\cite{zhang2000control,sikka2005tactile}, the projector matrix~\cite{li2013control,kappassov2020touch}, or other tactile-based controllers~\cite{she2021cable,wilson2023cable} to map tactile feedback onto robot actions. Learning-based tactile servoing leverages modern machine learning techniques, including imitation learning~\cite{sutanto2019learning} and deep learning~\cite{lepora2019pixels,lepora2021pose,lloyd2023pose}, to enable implicit mapping for complex robot manipulations. \method builds upon the fundamental principles of model-based tactile servoing, demonstrating its efficacy in the intricate manipulation of articulated objects.

\section{The \method Method}\label{sec:methods}

This section outlines our problem formulation for prior-free manipulation of articulated objects. At the core of \method is the ability to regulate stable contact throughout the manipulation process. This involves guiding the object from its initial to its final state using just tactile signals. We begin by introducing essential notations and preliminaries in \cref{sec:notation}, as these provide the foundational basis for our problem formulation and the subsequent derivations detailed in \cref{sec:formulation}. We then describe the contact representation in \cref{sec:representation}. This representation is crucial in determining whether a contact is stable and this is discussed in \cref{sec:contact}. Finally, \cref{sec:pose_calc} presents a computational method for calculating the robot's pose adjustments. These adjustments are essential for maintaining stable contact, thereby enabling the successful manipulation of articulated objects.

\subsection{Notation and Preliminaries}\label{sec:notation}

Before delving into the detailed formulation of \method, we start by defining the following notations:
\begin{itemize}[leftmargin=*,noitemsep,nolistsep]
    \item Bold lowercase letters represent vectors (\eg, \(\boldsymbol{p}\)), with the subscript \(i\) denoting the \(i\)-th entry of the vector (\eg, \(p_i\)).
    \item Bold uppercase letters represent matrices (\eg, \(\boldsymbol{M}\)). The identity matrix of dimension \(n\) is denoted by \(\boldsymbol{I}_{n \times n}\). The superscript \(\cdot^{\transpose}\) indicates the transpose.
    \item The symbol \(\hat{\cdot}\) indicates the axis of a frame. The generalized pose of a frame \(\{i\}\) relative to another frame \(\{j\}\) is described by a position vector \(\boldsymbol{p}_i^j \in \mathbb{R}^3\) and a rotation matrix \(\boldsymbol{R}_i^j \in SO(3)\). For simplicity, when frame \(\{j\}\) is equivalent to the world frame \(\{w\}\), the superscript is omitted. The homogeneous transformation matrix \(\boldsymbol{T}_i^j\in SE(3)\), integrating the position vector and rotation matrix, is given by:
    \begin{equation*}
        \boldsymbol{T}_i^j = \begin{bmatrix}
            \boldsymbol{R}_i^j & \boldsymbol{p}_i^j \\
            \begin{matrix}
                0 & 0 & 0
            \end{matrix} & 1
        \end{bmatrix}.
    \end{equation*}
    \item Calligraphic font letters, such as \(\mathcal{C}\), are used to denote sets. The notation \(\lvert\cdot\rvert\) represents the cardinality, or number of elements, in a set. Specially, \(\mathcal{K}_{ij}\) refers to a set of corresponding element pairs from sets \(\mathcal{C}_i\) and \(\mathcal{C}_j\), formulated as:
    \begin{equation*}
        \mathcal{K}_{ij} = \{(\boldsymbol{u}, \boldsymbol{v}) \mid \boldsymbol{u} \in \mathcal{C}_i, \boldsymbol{v} \in \mathcal{C}_j\}.
    \end{equation*}
\end{itemize}

\begin{figure}[t!]
    \centering
    \includegraphics[width=\linewidth]{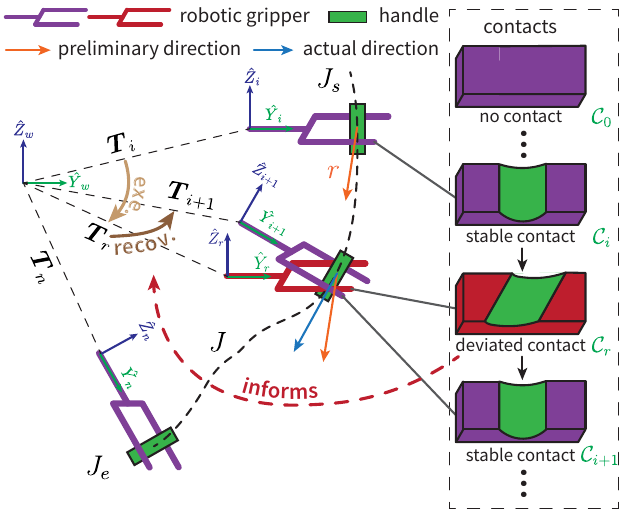}
    \caption{\textbf{Schematic of \method.} \method starts from an initial pose and an estimated preliminary direction. Thereafter, the pose is dynamically adjusted so as to maintain stable contact throughout the manipulation process. This continuous adjustment is crucial to ensure the correct interaction direction is also maintained. This culminates in the successful manipulation of the articulated object.}
    \label{fig:method}
\end{figure}

\subsection{Problem Formulation}\label{sec:formulation}

In this subsection, we formulate the process of tactile-informed prior-free manipulation of articulated objects. Consider an articulated object with a movable part, which is assumed to be graspable through a rigidly attached ``handle.'' We postulate that the object's articulation restricts the handle to a 1-\ac{dof} trajectory. Let us denote the handle's trajectory as \(J\), beginning at point \(J_s\) and concluding at \(J_e\). Initially, in the absence of priors, only \(J_s\) is known. A robotic gripper, positioned in a global pose \(\boldsymbol{T}_1\) is grasping the handle. The objective is to navigate the gripper through a series of poses, \(\boldsymbol{T}_1, \boldsymbol{T}_2, \boldsymbol{T}_3, \dots\), to maneuver the handle towards \(J_e\). At any given pose \(\boldsymbol{T}_i\), the contact state \(\mathcal{C}_i\) between the object handle and the robotic gripper can be ascertained via a mapping function \(f_c\):
\begin{equation}
    \mathcal{C}_i = f_c (J, \boldsymbol{T}_i).
    \label{eq:contact-def}
\end{equation}
In practical terms, \(\mathcal{C}_i\) is directly measurable through tactile sensors. Importantly, the state prior to establishing any contact is denoted as \(\mathcal{C}_{0}\).

\paragraph*{Stable contact}

To ensure successful manipulation, it is crucial that the contacts \(\mathcal{C}_{1:n}\) meet certain constraints, guaranteeing that the gripper can consistently grasp the handle throughout the process. In this paper, we characterize a state that meets these constraints as maintaining \textit{stable} contact.

\textit{Stable} contact is characterized by two essential features. First,  the gripper has to maintain a secure grip on the handle without causing damage to itself, while also preventing slippage. This ensures that any relative pose changes between the grip and handle are accurately mirrored in the contact changes. This concept is quantified by referencing state prior to establishing any contact \(\mathcal{C}_0\) and evaluating subsequent contact states \(\mathcal{C}_i\). We use the function \(f_d(\mathcal{C}_i, \mathcal{C}_0)\) to calculate the deformation of \(\mathcal{C}_i\); this deformation must remain within the elastic limits \(\boldsymbol{e}\) of the gripper material. Moreover, the shear deformation, as calculated by the function \(f_s(\mathcal{C}_i, \mathcal{C}_0)\), should not exceed \(s\), a threshold determined by the friction ratio between the handle and gripper material. Formally, this can be expressed as:
\begin{equation}
    \begin{bmatrix}
        f_e(\mathcal{C}_i, \mathcal{C}_0)\\
        f_s(\mathcal{C}_i,\mathcal{C}_0)
    \end{bmatrix}
    \leq
    \begin{bmatrix}
        \boldsymbol{e}\\
        s
    \end{bmatrix}.
    \label{eq:stable-contact}
\end{equation}

 Second, a \textit{stable} contact should guarantee that the movement of the gripper is synchronized with the handle's motion. To establish a benchmark for \textit{stable} contact \(\mathcal{C}_1\), an initial gripper pose \(\boldsymbol{T}_1\) is manually configured. This methodology is commonly employed in the manipulation of articulated objects~\cite{karayiannidis2016adaptive}. A contact \(\mathcal{C}_i\) is considered stable if its deviation from \(\mathcal{C}_1\) stays within a specified threshold \(d\):
\begin{equation}
    f_d(\mathcal{C}_i,\mathcal{C}_1) \leq d,
    \label{eq:effective-contact}
\end{equation}
where \(f_d(\mathcal{C}_i,\mathcal{C}_1)\) computes the difference between \(\mathcal{C}_i\) and \(\mathcal{C}_1\).

\paragraph*{Robot manipulation}

Upon establishing stable contact, the manipulation process commences with a preliminary direction relative to frame \(\{i\}\) for operating the articulated object. This direction, represented as \(r\), is encapsulated by a rotation matrix \(\boldsymbol{R}^{i} \in SO(3)\). A detailed discussion on obtaining it and its impact on the efficiency of the method is provided in Section \ref{sec:limitation}. Then, the homogeneous transformation matrix \(\boldsymbol{T}_{r}^{i}\), resulting from moving along \(r\) by a distance \(t\), is defined as:
\begin{equation}
   \boldsymbol{T}_{r}^{i} = \begin{bmatrix}
        \boldsymbol{I}_{3\times3} & \boldsymbol{R}^{i}\boldsymbol{t} \\
        \begin{matrix}
                0 & 0 & 0
            \end{matrix} & 1 
    \end{bmatrix},
    \label{eq:T-D}
\end{equation}
where \(\boldsymbol{t}\) denotes the translation vector whose Euclidean norm is equal to \(t\):
\begin{equation}
    \left\|\boldsymbol{t}\right\|_2 = t.
    \label{eq:translation_vector}
\end{equation}

The robot follows the preliminary direction, executing a series of gripper frame poses \(\boldsymbol{T}_{1:n}\) to guide the handle's transition from \(J_{s}\) to \(J_{e}\) while ensuring continuous stable contact. For each incremental step \(i = 1, \dots, n-1\), the next pose \(\boldsymbol{T}_{i+1}\) is computed as follows:
\begin{equation}
\begin{aligned}
 \boldsymbol{T}_{i+1} & = \boldsymbol{T}_{i} \boldsymbol{T}_{i+1}^{i}\\
    & = \boldsymbol{T}_{i} \underbracket{\boldsymbol{T}_r^{i}}_{\text{execution}} \underbracket{\argmin_{\boldsymbol{T}^{r}_{i+1}\in SE(3)} \left\Vert f_c(\boldsymbol{T}_i\boldsymbol{T}_r^{i}\boldsymbol{T}^{r}_{i+1}, J) - \mathcal{C}_1 \right\Vert_{2}}_{\text{recovery}}
\end{aligned}.
    \label{eq:formulation}
\end{equation}

\paragraph*{Execution}
The execution term in \cref{eq:formulation} involves proceeding to move along the preliminary direction as described in \cref{eq:T-D}.  This navigation progresses under the assumption of no prior knowledge about the object handle's trajectory, posing a risk of exceeding the tolerance limits necessary for maintaining stable contact. To balance this risk and maintain efficiency, it is crucial to carefully select the magnitude of \(t\) in \cref{eq:T-D}. The objective is to maximize \(t\) while ensuring that the constraints for stable contact are met. This optimization can be mathematically represented as:
\begin{equation}
    \begin{aligned}
        & \underset{t}{\text{maximize}}
        & & t \\
        & \text{subject to}
        & & f_e(f_c(\boldsymbol{T}_i\boldsymbol{T}_{r}^{i}, J),\mathcal{C}_0) \leq \boldsymbol{e},\\
        &&& f_s(f_c(\boldsymbol{T}_i\boldsymbol{T}_{r}^{i}, J),\mathcal{C}_0) \leq s, \\
        &&& f_d(f_c(\boldsymbol{T}_i\boldsymbol{T}_{r}^{i}, J),\mathcal{C}_1) \leq d.
    \end{aligned}
    \label{eq:y}
\end{equation}

\paragraph*{Recovery}

The recovery term in \cref{eq:formulation} implies that at each step, the robot adjusts its pose to minimize any deviation from the stable contact state \(\mathcal{C}_1\), which may arise due to the execution step. 

\subsection{Contact Representation}\label{sec:representation}

For the effective computation of \cref{eq:formulation} and optimization of \cref{eq:y}, a suitable representation of \(\mathcal{C}_i\) is crucial. To ensure compatibility with the homogeneous transformation matrix and to match the capabilities of current tactile sensors, we represent the contact \(\mathcal{C}_i\) as a set of positions of all discretized points at the contact site, relative to the reference frame \(\{i\}\). The position of each contact point within \(\mathcal{C}_i\) is denoted by \(\boldsymbol{p}^i\). To facilitate alignment with the homogeneous transformation matrix, the column vector representation of \(\boldsymbol{p}^i\) is defined in an augmented format as follows:
\begin{equation}
    \boldsymbol{p}^i =
    \begin{bmatrix}
        p_x^i &
        p_y^i &
        p_z^i &
        1
    \end{bmatrix}^{\transpose},
\end{equation}
where \(p_x^i\), \(p_y^i\), and \(p_z^i\) represent the coordinates relative to the gripper frame \(\{i\}\).

Significantly, \(\mathcal{C}_0\) includes all points on the gripper's outer surface. Assuming the total number of points is \(m\), \(\mathcal{C}_0\) is represented as:

\begin{equation}
   \mathcal{C}_0 = \{\boldsymbol{p}_j^0 \mid j = 1, \dots, m\}.
\end{equation}

For simplicity, we assume a flat gripper surface, a common assumption in current robotic grippers and easily achievable with tactile sensor fabrication. Consequently, \(p_x^0\) is consistent across all points and set to \(0\), attainable through coordinate transformation. The position of a contact point at \(\mathcal{C}_i\) is then characterized by the condition: 
\begin{equation}
    \vert p^i_{1}\vert \geq \epsilon
    \label{eq:contact_point}
\end{equation}
where \(\epsilon\) is a small positive number, indicating that deformation in the \(\hat{x}\) direction must exceed \(\epsilon\) for it to be considered contact. 

Therefore, \(\mathcal{C}_i\) for \(i = 1 \dots n\) can be defined as:
\begin{equation}
    \mathcal{C}_i = \{\boldsymbol{p}^i \mid \vert p^i_1\vert \geq \epsilon \}.
\end{equation}

\subsection{Stable Contact Formulation}\label{sec:contact}

With the established representation of contact, we now turn to the computational process for determining stable contact.

The computation of stable contact involves addressing the gripper's maximum elastic deformation, encompassing both normal and shear components. These are captured in the expressions for \(f_e(\mathcal{C}_i,\mathcal{C}_0)\) and \(\boldsymbol{e}\) within the context of stable contact. Ensuring that the contact deformation remains within the material's elastic limits is crucial for maintaining a stable grip on the object:
\begin{equation}%
    f_e(\mathcal{C}_i,\mathcal{C}_0) = 
    \begin{bmatrix}
        \max\{\vert u_{1}\vert \mid \boldsymbol{u}\in \mathcal{C}_i\} \\
        \max\left\{\left\|
        \begin{bmatrix}
            u_{2}-v_{2}\\
            u_{3}-v_{3}
        \end{bmatrix}
        \right\|_{2} \; \middle| \; (\boldsymbol{u},\boldsymbol{v})\in\mathcal{K}_{0i}\right\} 
    \end{bmatrix} 
\end{equation}
and
\begin{equation}
    \boldsymbol{e} = \begin{bmatrix}
        e_n &
        e_s
    \end{bmatrix}^{\transpose},
    \label{eq:c_n&c_S}
\end{equation}
where \(d_n\) and \(d_s\) denote the maximum normal and shear elastic deformation, respectively. For notation simplicity, we denote \(\vert u_{1}\vert\) as \(\Delta N\) and \(\displaystyle\left\|%
\begin{bmatrix}
    u_{2}-v_{2}\\
    u_{3}-v_{3}
\end{bmatrix}%
\right\|_{2}\) as \(\Delta S\) henceforth.

Similarly, we formulate \(f_s(\mathcal{C}_i,\mathcal{C}_0)\) and \(s\) to address concerns related to slipping in \cref{eq:stable-contact}. Practically, to prevent sliding, the friction \(f\) between the object and the gripper should not surpass the maximum static friction \(f_{M}\). In our model, the friction at \(\mathcal{C}_i\) is positively correlated with the shear deformation observed from \(\mathcal{C}_0\):
\begin{equation}
    f \sim \Delta S.
\end{equation}

Thus, the shear deformation must adhere to the following constraints:
\begin{equation}
    \Delta S  \leq \delta^M(f_N), \forall (\boldsymbol{u},\boldsymbol{v})\in\mathcal{K}_{0i},
\end{equation}
where \(\delta^M\) represents the deformation associated with the maximum static friction \(f_{M}\), proportional to the normal force \(f_N\). Assuming linear elasticity in the normal direction, \(\delta^M\) can be modeled as:
\begin{equation}
     \delta^M(\Delta N) = \delta_0\Delta N, (\boldsymbol{u},\boldsymbol{v})\in\mathcal{K}_{0i},
    \label{eq:delta}
\end{equation}
with \(\delta_0\) being a constant determined experimentally:
\begin{equation}
      \delta_0 = \frac{\delta^M(\Delta N_0)}{\Delta N_0}, (\boldsymbol{u},\boldsymbol{v})\in\mathcal{K}_{0i}.
    \label{eq:delt_0}
\end{equation}

The no-sliding constraint is thus defined as:
\begin{equation}
     \frac{\Delta S}{\Delta N}   \leq \delta_0, \forall (\boldsymbol{u},\boldsymbol{v})\in\mathcal{K}_{0i},
\end{equation}
leading to the expression for \(f_s(\mathcal{C}_i,\mathcal{C}_0)\):
\begin{equation}
    f_s(\mathcal{C}_i,\mathcal{C}_0) = \max\left\{\frac{\Delta S}{\Delta N} \; \middle| \; (\boldsymbol{u},\boldsymbol{v})\in\mathcal{K}_{0i}\right\},   
\end{equation}
and setting
\begin{equation}
    s = \delta_0.
    \label{eq:delta-0}
\end{equation}

Additionally, in the context of \cref{eq:effective-contact}, we intuitively compute it as the difference between the corresponding points of two contacts, considering our established contact representation. To quantify this difference, we apply the loss function outlined in Besl \etal~\cite{besl1992method}. However, to ensure fairness in cases where a contact has a larger number of contact points, we normalize this loss by the number of corresponding pairs in \(\mathcal{K}_{1i}\):
\begin{equation}
    f_d(\mathcal{C}_i,\mathcal{C}_1) = \frac{\sum_{(\boldsymbol{u},\boldsymbol{v}) \in \mathcal{K}_{1i}} \|\boldsymbol{u} - \boldsymbol{v}\|_{2}}{\vert\mathcal{K}_{1i}\vert}.
    \label{eq:d}
\end{equation}

\subsection{Robot Pose Update}\label{sec:pose_calc}

Having established the contact representation and computation methods for stable contact constraints, we should address the optimization in \cref{eq:y} to compute the value of \(t\) for the determination of \(\boldsymbol{T}_{r}^{i}\). However, in the absence of knowledge about the handle's trajectory \(J\), determining \(t\) explicitly remains challenging. Therefore, we adopt an exploration mechanism by continuously executing with small increments in \(t\) until any term in the constraints of optimization \cref{eq:y} reaches its upper bound, scaled by a safety margin \(\alpha\) (\(0<\alpha<1\)). This process is concluded as
\begin{equation}
    \boldsymbol{T}_{r} = \boldsymbol{T}_i\boldsymbol{T}_{r}^{i},
\end{equation}
which is referred to as the \textit{execution stage}.

To recover the contact to within the specified bounds after the execution stage, we employ the Kabsch algorithm~\cite{kabsch1976solution} to compute the optimal transformation \(\boldsymbol{T}^{r}_{i+1}\) for the recovery term in \cref{eq:formulation}:
\begin{equation}
    \boldsymbol{T}^{r}_{i+1} = \argmin_{\boldsymbol{T}^{r}_{i+1}\in SE(3)}\sum_{(\boldsymbol{u},\boldsymbol{v})\in\mathcal{K}_{1r}}\|\boldsymbol{T}^{r}_{i+1}\boldsymbol{u}-\boldsymbol{v}\|_{2}.
    \label{eq:stage2}
\end{equation}

The updated robot pose is then given by:
\begin{equation}
    \boldsymbol{T}_{i+1} = \boldsymbol{T}_{r}\boldsymbol{T}^{r}_{i+1}.
\end{equation}
This phase is referred to as the \textit{recovery stage}. The update of the robot pose follows an iteration between the two stages. 

\begin{figure}[t!]
    \centering
    \includegraphics[width=\linewidth]{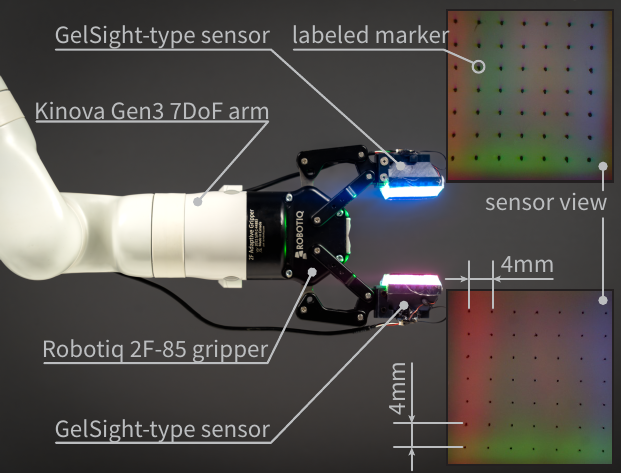}
    \caption{\textbf{Real-world experiment setup.} The experiments are conducted in a real-world environment using a Gen3 7DoF robotic arm paired with a Robotiq 2F-85 gripper. To equip it with the ability to sense contact geometry, we adopt two GelSight-type tactile sensors, each featuring a \(7 \times 7\) grid of labeled markers for point correspondence between frames.}
    \label{fig:exp-setup}
\end{figure}

\section{Real-world Experiments}\label{sec:real-world_exp}

To assess the effectiveness of \method, our proposed approach to prior-free manipulation of articulated objects, we conducted a series of real-world experiments. These were designed with meticulous attention to detail, as outlined in \cref{sec:real-world_exp_setup}. They aim to thoroughly evaluate key constraints and determine essential hyper-parameters for the task, as discussed in \cref{sec:methods}. Our experiments showcase the strengths of \method, particularly its robustness and adaptability under various conditions. These include scenarios where priors for object articulation are ambiguous (\cref{sec:under-ambiguity}), imperfect (\cref{sec:under-uncertainties}), unknown (\cref{sec:intricate_traj}), or obsolescent (\cref{sec:under-perturbation}), as previously discussed in \cref{sec:intro}.

\subsection{Experimental Setup}\label{sec:real-world_exp_setup}

\begin{figure}[t!]
    \centering
    \includegraphics[width=\linewidth]{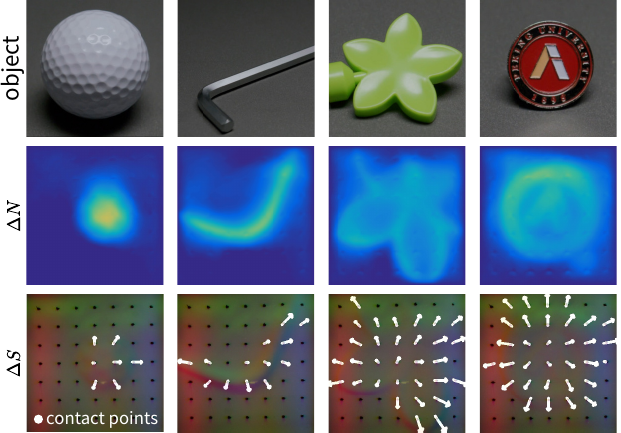}
    \caption{\textbf{Visualization of the contact geometry detected by the sensors adopted.} The GelSight-type sensors are adept at providing contact geometry information \(\Delta N\) in the normal direction during object interactions. Labeled markers are used for accurate point correspondence to further enhance their effectiveness. Utilizing these features, \(\Delta S\) is calculated and visualized using white arrows in the illustration. For better visibility, the magnitude of these arrows is exaggerated by a factor of \(5\), ensuring a clearer visualization of the contact dynamics.}
    \label{fig:contact_vis}
\end{figure}

\paragraph*{Robot system setup}

\method is designed for broad compatibility with robotic systems capable of facilitating 6D movement in the gripper. In experiments, we utilized a Kinova Gen3 7DoF robotic arm equipped with a Robotiq 2F-85 gripper (\cref{fig:exp-setup}). To assess the contact constraints as outlined in \cref{sec:methods}, we integrated two GelSight-type tactile sensors into the gripper.

\paragraph*{Tactile sensor}

To further enhance our ability to validate these constraints, we adopt GelSight-type tactile sensors, which effectively provide geometry information at the contact surface. The back of the deformable pad is rigidly attached to a rigid acrylic sheet. Each sensor was specially fabricated with a unique \(7\times7\) grid of labeled markers (\cref{fig:contact_vis}) on the front of the pad. Without any surrounding structural geometry, these markers share the same mechanical characteristics, allowing for calibration using consistent parameters. This grid aids in accurately identifying point correspondences between two contacts. Although this modification slightly reduces the resolution and depth reconstruction accuracy as compared to conventional GelSight sensors, its effectiveness in practical scenarios and in validating the contact constraints detailed in \cref{sec:methods} is notable.

\begin{figure}[t!]
    \centering
    \includegraphics[width=\linewidth]{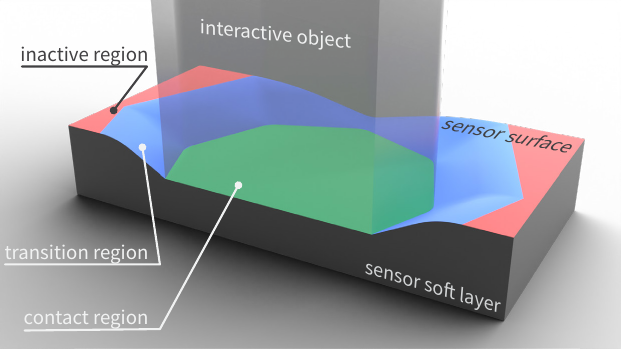}
    \caption{\textbf{Selection concerns for the \(\epsilon\) value.} The compliance inherent in the tactile sensor material results in the activation of both the direct contact region and the adjacent transitional areas when contacting the handle. This characteristic underlines the need for a carefully adaptable \(\epsilon\) value. Such adaptability ensures the precise identification of true contact points, thereby reducing the likelihood of detecting unreliable points.}
    \label{fig:real_contact}
\end{figure}

\paragraph*{System hyper-parameters}

\begin{figure}[t!]
    \centering
    \includegraphics[width=\linewidth]{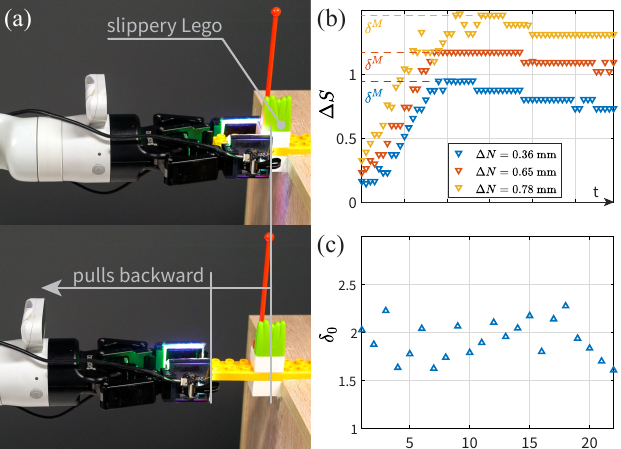}
    \caption{\textbf{\(\boldsymbol{\delta_0}\) value acquisition.} The experiment to obtain hyper-parameter \(\delta_0\), as defined in \cref{eq:delt_0}: (a) The experiment begins with the arm pulling backward after the gripper has firmly grasped a slippery Lego piece. (b) Multiple trials are conducted at different values of \(\Delta N\). Three representative trials are visualized to illustrate the process. The results follow the friction model well and \(\delta^M\) can be easily identified.  The results align well with the friction model, allowing for the straightforward identification of \(\delta^M\). (c) The \(\delta_0\) obtained from these multiple trials is visualized, and a value of \(1.5\) is selected for safety considerations.}
    \label{fig:slip_test}
\end{figure}

The determination of values for the defined parameters is crucial. The following outlines the chosen values, with a summary provided in \cref{tab:hyper-params}:
\begin{itemize}[leftmargin=*,noitemsep,nolistsep]
    \item The threshold for identifying contact points, \(\epsilon\) in \cref{eq:contact_point}, is set to be adaptable. This is due to the material's flexibility, as shown in \cref{fig:real_contact}, which results in deformation not only in the contact region but also in the surrounding area upon contact with the object. This deformation relates to \(\Delta N\) and the handle's shape, among other factors. For adaptability, \(\epsilon\) is defined as the minimal \(\Delta N\) that ensures at least 8 contact points, exceeding the number of markers in a line (7 in our sensors), thus ensuring these points are non-colinear.

    \item The normal elastic deformation limit, \(e_n\) in \cref{eq:c_n&c_S}, is set at \(2~\mathrm{mm}\). This value is chosen based on observations that within this range, the system can revert to its initial state after releasing contact. It is important to note that this value represents a conservatively reduced normal elastic deformation for enforcing stricter constraints.
    
    \item The parameter \(\delta_0\) in \cref{eq:delt_0} is linked to handle material properties. To set a lower bound, an experiment with a slippery Lego piece, depicted in \cref{fig:slip_test}(a), identifies the \(\delta^M\) values at different \(\Delta N\). This is done by tracking the peak shear displacement for each marker, as shown in \cref{fig:slip_test}(b). The results from multiple trials, illustrated in \cref{fig:slip_test}(c), lead to setting \(\delta_0\) at \(1.5\) for safety.

    \item The shear elastic deformation limit, \(e_s\) in \cref{eq:c_n&c_S}, is calculated as \(e_n\delta_0\). Though smaller than the actual one, this value is deemed acceptable for stricter constraint enforcement.
    
    \item The upper bound for another stable contact constraint, \(d\) in \cref{eq:effective-contact}, is experimentally determined to be \(0.4~\mathrm{mm}\), ensuring it does not exceed \(e_n\), \(e_s\), and \(\delta_0\) during manipulation.

    \item The movement magnitude in the \textit{execution stage}, the small increment in \(t\) established in \cref{eq:y}, is set as \(\epsilon \delta_0\) to preserve stable contact during execution.
\end{itemize}

\begin{table}[ht!]
    \centering
    \small
    \setlength{\tabcolsep}{3pt}
    \caption{\textbf{Values of system hyper-parameters}}
    \label{tab:hyper-params}
    \resizebox{\linewidth}{!}{%
        \begin{tabular}{ccccccc}
            \toprule
            \textbf{hyper-parameter} & \(\epsilon\) & \(e_n\) & \(e_s\) & \(\delta_0\) & \(d\) & \(\alpha\)\\
            \midrule
            \textbf{value} & adaptive & \(2~\mathrm{mm}\) & \(3~\mathrm{mm}\) & \(1.5\) & \(0.4~\mathrm{mm}\) & \(0.6\)\\
            \bottomrule
        \end{tabular}%
    }%
\end{table}

The experiments described below are conducted in accordance with the setup and hyper-parameters detailed above. This ensures consistency and relevance in evaluating the efficacy of \method in real-world scenarios.

\subsection{Manipulation under Ambiguous Priors}\label{sec:under-ambiguity}

\begin{figure*}[t!]
    \centering
    \includegraphics[width=\linewidth]{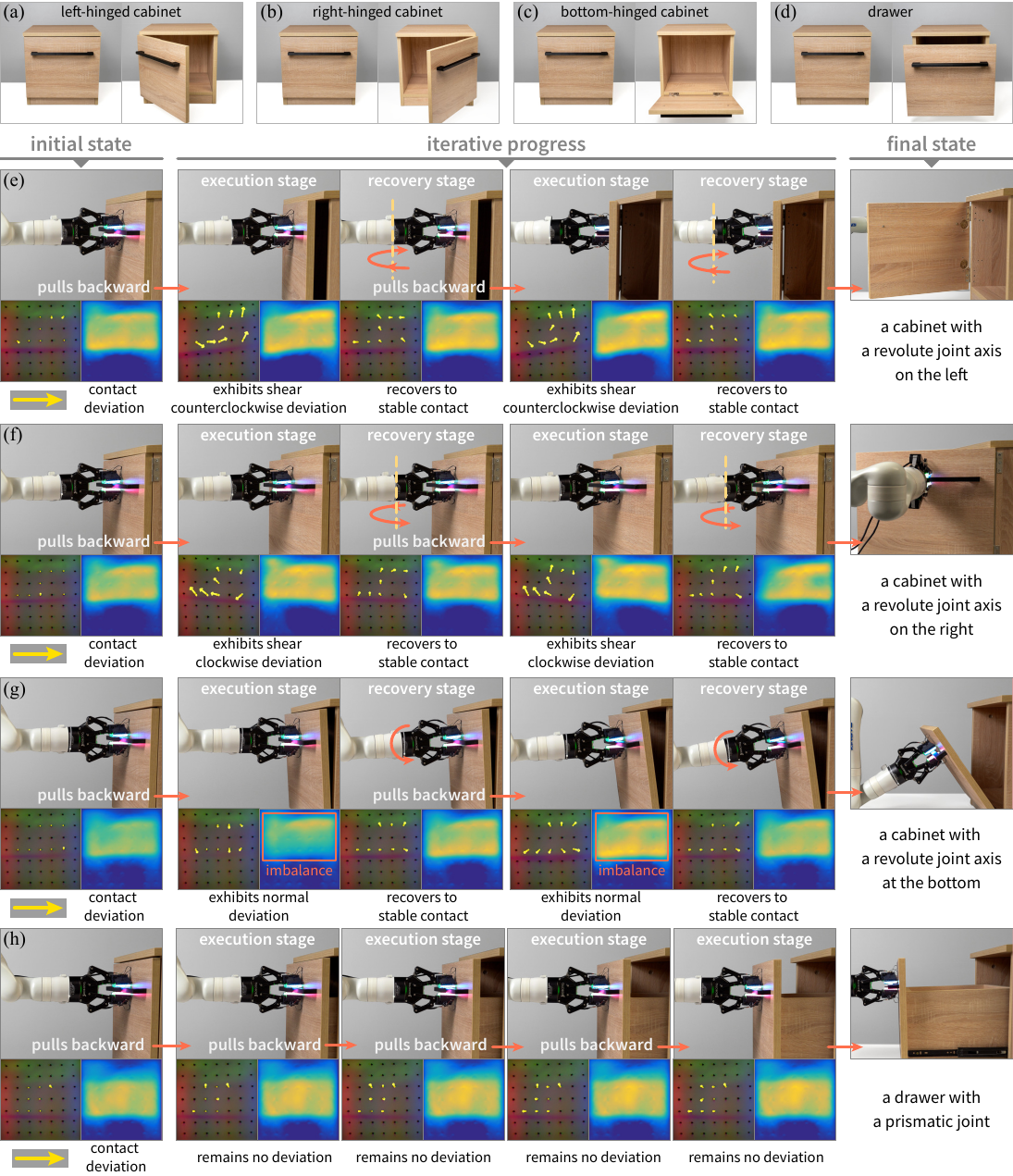}
    \caption{\textbf{Manipulation under ambiguous priors.} \method showcases its proficiency in addressing the challenge caused by ambiguous priors, characterized by objects that, while visually indistinguishable, feature distinct articulation mechanisms. Our evaluation involves four articulated objects that look identical but differ in articulation: (a) a cabinet with a revolute joint on the left; (b) another cabinet with a revolute joint on the right; (c) a cabinet with a revolute joint at the bottom; and (d) a drawer with a prismatic joint. In scenarios (e)--(h), corresponding to objects (a)--(d), \method starts from the same position and in the same direction. The initial backward pull by the arm reveals variations in contact loss due to the differing articulations. Upon reaching a certain deviation threshold, the system triggers an adjustment in the arm's pose. This adjustment, optimized by \cref{eq:stage2}, aims to recover stable contact and re-align to the correct interaction direction. The arm then continues its backward pull. Through this iterative process of adjustment and execution, each object, starting from the same initial state, is maneuvered to a distinct final state reflective of its unique articulation. For  video demonstrations, refer to \href{\suppUrlSI}{Supp Video S1}.}
    \label{fig:under-ambiguity}
\end{figure*}

Ambiguous visual priors pose a significant challenge in robotic manipulation, especially when objects with visually indistinguishable appearances harbor varying articulation mechanisms. This dilemma is discussed in \cref{sec:rw:explicit-prior} and exemplified by Zhu \etal~\cite{zhu2020dark}. To illustrate, \cref{fig:under-ambiguity}(a)--(d) features four objects that, despite their identical appearances, are equipped with different articulations: a cabinet with a revolute joint on the left (a), another with the joint on the right (b), a third with a joint at the bottom (c), and a drawer with a prismatic joint (d). This diversity underscores the complexity of inferring object kinematics solely from visual information.

\begin{figure*}[t!]
    \centering
    \includegraphics[width=\linewidth]{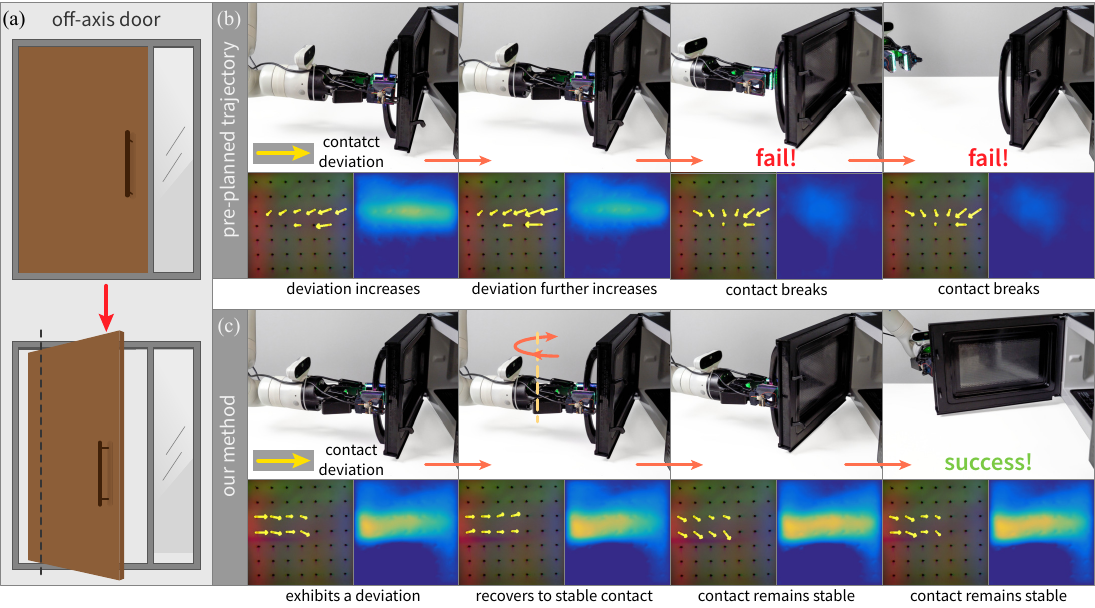}
    \caption{\textbf{Manipulation under imperfect priors.} \method adeptly addresses challenges posed by imperfect priors, which accurately identify the model type but contain erroneous parameters. (a) An example is an off-axis door, designed for space and energy efficiency, diverging from traditional models by not placing the hinge directly on one side. This variation can mislead prior assumptions regarding the arc trajectory's radius. (b) We mirror this scenario by assigning the robot a pre-programmed arc trajectory to open a microwave oven with a revolute joint, incorporating a \(10~\%\) radius error. This slight misalignment between the priors and actual parameters leads to a gradual loss, eventually causing the gripper to lose contact with the handle, resulting in a failed manipulation attempt. (c) In contrast, \method dynamically re-calibrates the arm's pose to correct any unexpected errors, ensuring alignment with the correct interaction direction, going on to successfully open the microwave oven. This demonstration of adaptability underscores \method's ability to complete tasks even with imperfect priors. For video demonstrations, refer to \href{\suppUrlSII}{Supp Video S2}.}
    \label{fig:uncertain_parameter}
\end{figure*}

\method effectively demonstrates proficient manipulation of distinct objects in all four cases, as shown in \cref{fig:under-ambiguity}(e)--(h), each corresponding to the configurations in \cref{fig:under-ambiguity}(a)--(d). Starting from the same initial state with stable contact \(\mathcal{C}_1\), \method employs an iterative two-stage cycle, beginning with an identical preliminary direction \(\boldsymbol{R}^{1}\). During the \textit{execution stage}, tactile feedback is utilized to discern deviations from the correct interaction direction by observing changes in the contact points. For example, the contact in (e) exhibits a shear counterclockwise deviation, indicating a need for an additional clockwise rotation in the robot's movement, and conversely for the shear clockwise deviation observed in (f). In the scenario depicted in (g), a deviation in the normal direction \(\Delta N\) signals the requirement for an added rotation for recovery. Conversely, in (h), the contact remains stable as the arm's movement is congruent with the correct direction. When a predetermined deviation threshold is reached, the system enters the \textit{recovery stage}, employing the optimization outlined in \cref{eq:stage2} to compute the optimal arm pose that compensates for the observed deviation, thereby realigning the gripper to the correct interaction direction. Following this adjustment, the process resumes along the preliminary direction \(\boldsymbol{R}^{i}\). Through continual refinement to preserve stable contact, the arm successfully manipulates the objects into their intended states, as depicted in the last images in \cref{fig:under-ambiguity}.

\begin{figure*}[t!]
    \centering
    \includegraphics[width=\linewidth]{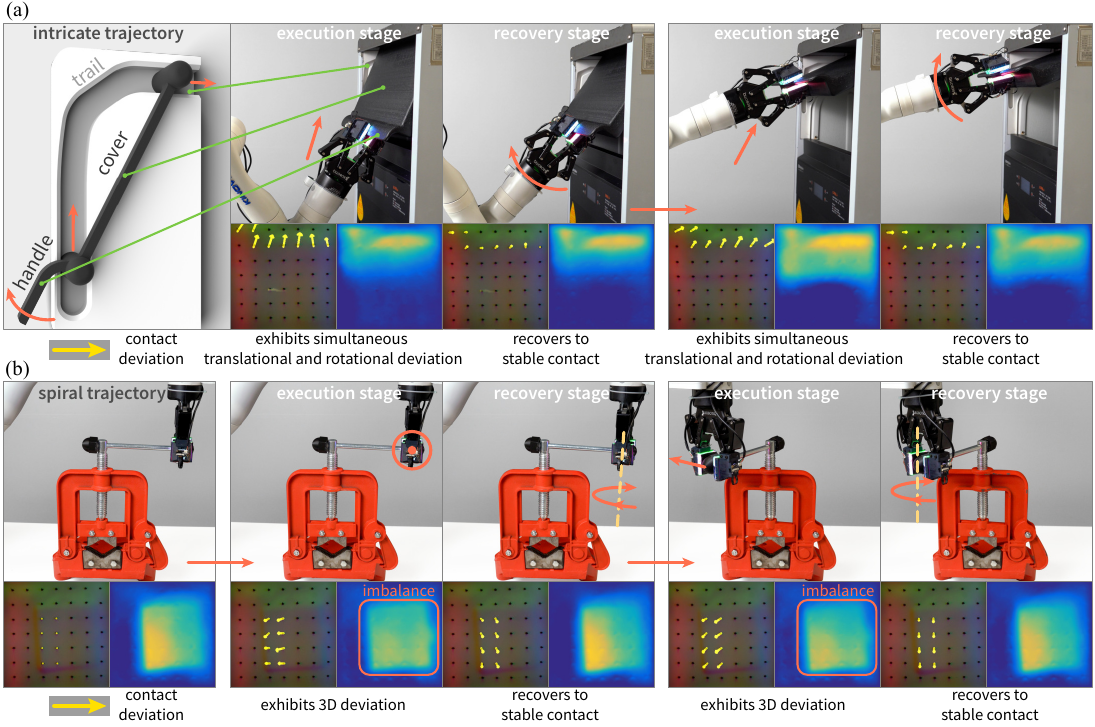}
    \caption{\textbf{Manipulation under unknown priors.} \method showcases exceptional proficiency in managing scenarios characterized by unknown priors, particularly when object kinematics defy prior-based prediction or exceed the limits of current robotic perception. We illustrate this capability with two distinct examples. In scenario (a), the object's articulation mechanism, which is concealed beneath a cover, involves motions that combine translation and rotation---a complexity not easily captured by standard prismatic or revolute joint models. Scenario (b) presents a handle executing a helical trajectory, with the precise determination of its pitch posing a substantial challenge. Through these examples, \method proves its adeptness at navigating the intricacies of complex kinematic patterns, underscoring its effectiveness in scenarios where traditional perception models may falter. For video demonstrations, refer to \href{\suppUrlSIII}{Supp Video S3}.}
    \label{fig:intricate_traj}
\end{figure*}

\subsection{Manipulation under Imperfect Priors}\label{sec:under-uncertainties}

Imperfect priors present a significant challenge in robotic manipulation, even when the articulation structure is known. Errors in model parameters can make these priors unreliable, as demonstrated by the example of an off-axis door, depicted in \cref{fig:uncertain_parameter}(a). Unlike conventional doors with hinges located along one edge, the off-axis door features a hinge positioned away from the edge, complicating the acquisition of accurate priors.

The impact of imperfect priors is illustrated through a challenging manipulation task, in which the robot is programmed to follow an arc trajectory to open a microwave oven with a revolute joint, as shown in \cref{fig:uncertain_parameter}(b). However, a \(10~\%\) error in the trajectory's radius introduces a small but critical imperfection in the priors. This error, without any real-time corrective adjustments, accumulates a loss that eventually leads to the gripper losing contact with the handle, resulting in task failure.

In contrast, \cref{fig:uncertain_parameter}(c) demonstrates \method's approach, employing adaptive adjustments during the manipulation to effectively counteract the inaccuracies in the priors. By dynamically compensating for the unforeseen deviation, \method ensures alignment with the correct interaction direction. This adaptability allows \method to successfully complete the task, fully opening the microwave oven door, showcasing its efficacy in overcoming the limitations imposed by imperfect priors.

\subsection{Manipulation under Unknown Priors}\label{sec:intricate_traj}

Encountering objects whose articulation mechanisms defy accurate perception introduces a scenario where priors on kinematics remain unknown. Such conditions, characterized by complex movements that cannot be succinctly represented by conventional kinematic models like prismatic or revolute joints, present a significant challenge. Existing research has scarcely addressed the manipulation of objects requiring combined translational and rotational motions due to the difficulty in modeling these actions explicitly. For example, \cref{fig:intricate_traj}(a) illustrates an object designed compactly, necessitating that its handle executes both linear and rotational movement simultaneously. Another example is shown in \cref{fig:intricate_traj}(b), where a vise handle follows a spiral path. The ability to accurately identify and model such complex motion patterns remains beyond the reach of current perception methods.

\method circumvents this challenge, as demonstrated with two examples of objects whose handles exhibit complex movements combining translational and rotational motions. Through iterative processing, the \method adjusts the robot to the correct interaction direction and successfully manipulates the object in both examples.

\begin{figure*}[t!]
    \centering
    \includegraphics[width=\linewidth]{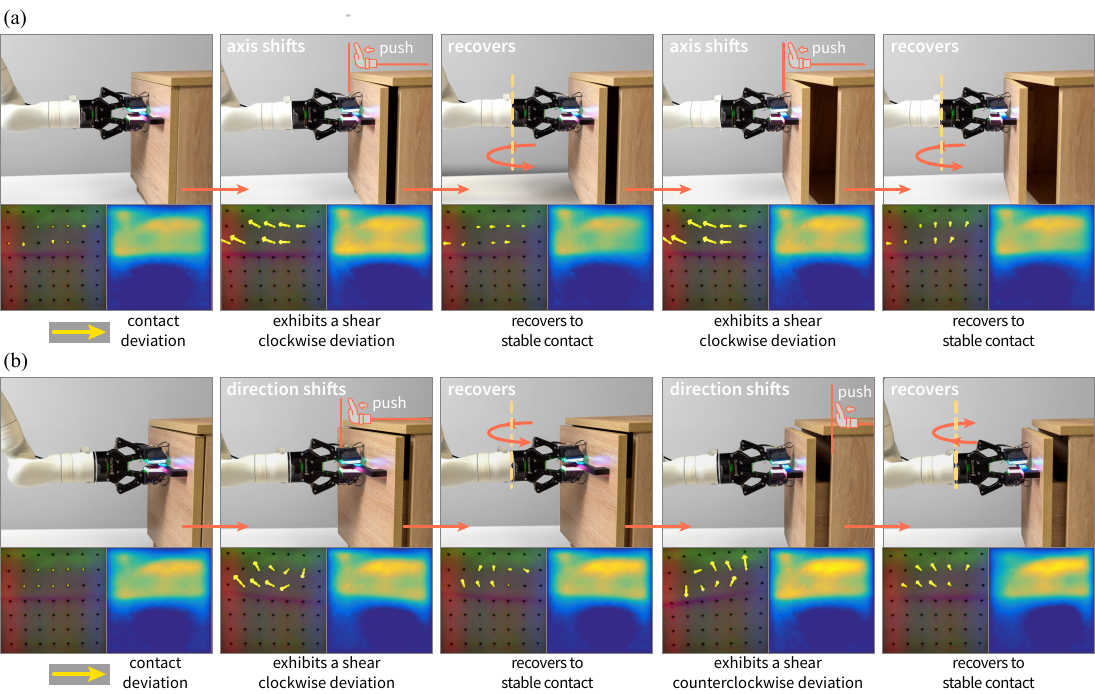}
    \caption{\textbf{Manipulation under obsolescent priors.} \method demonstrates exceptional performance in scenarios characterized by obsolescent priors, often resulting from unpredictable perturbations that modify object kinematics. Such perturbations are common in dynamic, human-centric environments, frequently emerging from unanticipated human interactions. To assess \method's resilience against these perturbations, we introduced random forces to one side of the object in scenarios (a) and (b), effectively altering the cabinet's rotation axis in (a) and the drawer's opening direction in (b). Despite the introduction of errors through these human-induced disturbances, \method adeptly adjusts, preserving the correct interaction direction and ensuring successful manipulation outcomes. For video demonstrations, refer to \href{\suppUrlSIV}{Supp Video S4}.}
    \label{fig:perturbation}
\end{figure*}

\subsection{Manipulation under Obsolescent Priors}\label{sec:under-perturbation}

Dynamic environments often introduce unpredictable perturbations, such as human interventions or environmental turbulence, rendering initially accurate priors for object kinematics obsolescent. To evaluate \method's performance in this context, we conducted experiments depicted in \cref{fig:perturbation} in which the position of the articulated object was deliberately altered during manipulation. This was achieved by applying force to one side of the object, leading to a shift in either the rotation axis of a cabinet (\cref{fig:perturbation}(a)) or the opening direction of a drawer (\cref{fig:perturbation}(b)), as indicated by the hand icons.

Despite encountering such unpredictable disturbances, \method demonstrates remarkable adaptability. It swiftly compensates for the changes by correctly adjusting the robot's interaction strategy, thus enabling successful manipulation. This resilience underscores \method's ability to effectively handle real-time perturbations, ensuring consistent manipulation performance even when initial kinematic priors are no longer valid due to unforeseen environmental interactions.

\section{Large-scale Verification in Simulation}\label{sec:simulation}

Given the logistical challenges of physically experimenting with a comprehensive variety of objects, we look to simulation studies to further validate the generalization capability of \method. Utilizing NVIDIA Isaac Sim, our simulations cover a wide range of objects, mirroring the setup of our real-world experiments as outlined in \cref{sec:sim_setup}. For a thorough assessment, we incorporate objects with prismatic and revolute joints from the PartNet-Mobility dataset~\cite{xiang2020sapien}, reflecting the common articulations found in existing datasets.

Acknowledging the limitation of current datasets of objects with simpler joint types, we further extend our verification to include articulated objects designed with complex manipulation trajectories. These objects are uniquely characterized by trajectories that follow randomly sampled curves, introducing additional challenges to the manipulation task. 

 To showcase the effectiveness of \method, we select several representative prior-based and prior-free methods as baselines, as detailed in \cref{sec:baseline}. In \cref{sec:sim-results}, we present both qualitative and quantitative results from our simulations. The findings compellingly demonstrate \method's proficiency in manipulating objects with a wide range of articulations, highlighting its adaptability and superior generalization across diverse scenarios compared to the baseline methods.

\subsection{Environmental Setup}\label{sec:sim_setup}

\begin{figure*}[t!]
    \centering
    \includegraphics[width=\linewidth]{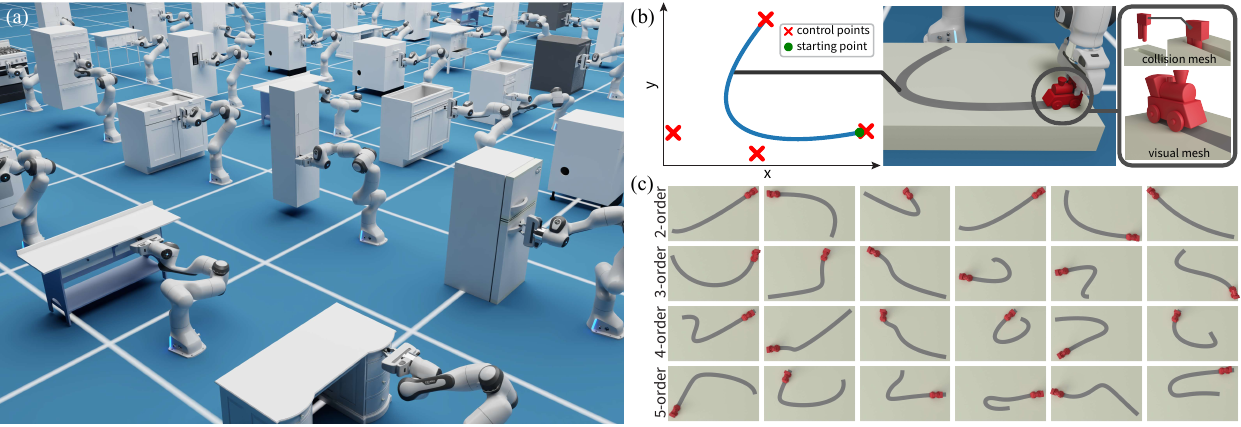}
    \caption{\textbf{Simulation setup.} (a) The setup includes objects with prismatic and revolute joints from the PartNet-Mobility dataset~\cite{mo2019partnet}, shown in their initial simulation states. (b) We further introduce a playboard featuring an intricate manipulation trajectory based on a randomly sampled Bézier curve; control points of the curve are marked with red ``x''s. At the starting point of this curve, a toy train is placed to act as the manipulation handle (right), with a detailed collision setup (below). (c) Further examples of generated playboards, where rows 1-4 correspond to playboards with 2nd, 3rd, 4th, and 5th-order Bézier curves as the trajectories, showcasing the range of complexity in manipulation paths.}
    \label{fig:sim-setup}
\end{figure*}

Our simulation environment, depicted in \cref{fig:sim-setup}, features a robot arm equipped with a gripper poised to engage an object's handle. Initially, the robot is programmed to follow a preliminary direction \(\boldsymbol{R}^1\), applying \method to articulate the object. A critical aspect of \method is its reliance on the compliance of silica gel material to accurately track contact between the gripper and the handle. To mimic this in a rigid body simulation environment, we employ specialized designs for contact measurement, detailed in \cref{appendix:sim-details:setup}.

The simulation divides test objects into two categories: standard objects with prismatic or revolute joints featuring various handle designs from existing datasets, and randomly created custom objects with complex trajectories. From the PartNet-Mobility dataset~\cite{xiang2020sapien}, we select objects with detailed articulation annotations provided by GAPartNet~\cite{geng2023gapartnet}, as outlined in \cref{appendix:sim-details:gapartnet-preproc}. For the evaluation, \(25\) objects with prismatic joints and another \(25\) with revolute joints were pre-processed. A manipulation is deemed successful if a revolute joint rotates beyond \(60.0^\circ\)or a prismatic joint extends over \(250.0~\mathrm{mm}\). \cref{fig:sim-setup}(a) showcases a subset of these objects and their initial grasps.

To assess \method's adaptability beyond basic joint types, we also evaluate it on articulated objects designed with randomly generated trajectories. These are constructed by sampling Bézier curve segments of various orders (\(2\), \(3\), \(4\), and \(5\)) on a \(2\)D plane, ensuring no self-intersections. Each curve inspires a \(3\)D playboard that defines a manipulation path, with a toy train at the start point acting as the handle. Physical collisions are configured to restrict the handle's movement along the trajectory, with the construction process detailed in \cref{appendix:sim-details:playboard}. \cref{fig:sim-setup}(b) illustrates an example of a \(3\)-order Bézier curve, its corresponding play-board, and the simulation's initial state. Additional examples of generated play-boards are presented in \cref{fig:sim-setup}(c), highlighting the increased complexity and manipulation challenge with higher-order curves. For the evaluation, \(25\) curves of orders \(2\) to \(5\) are sampled, each tested once. Success is achieved when the gripper moves the toy train from one end of the curve to the other.

\subsection{ Selected Baselines}\label{sec:baseline}

 To demonstrate the effectiveness of \method, we also select several representative prior-based or prior-free methods designed for the manipulation of articulated objects to conduct comparisons with \method. Specifically, we evaluate the performance of three distinct techniques: the prior-based pre-planned trajectory generation method, impedance control in a prior-free setting, and a vision-based learning approach, FlowBot3D~\cite{eisner2022flowbot3d}. Below, we detail the setup for each of these methods:

\paragraph*{Prior-based pre-planned trajectory}
 This method is highly effective when the kinematic priors of articulated objects are precisely known. However, meeting this requirement proves challenging in the real world due to pervasive uncertainties and unexpected perturbations~\cite{zhu2020dark}. To account for this in our simulations, we introduce noise \(\xi\) into the ideal kinematic model of the test objects. This noise \(\xi\) is modeled as a normal distribution with a mean of zero and a variance of \(0.01\). We explore the impact of this noise on the pre-planned trajectory generation method under the same initial conditions as \method for different joint types. For revolute joints, the relationship between the actual handle trajectory radius \(r_{gt}\) and the manipulated trajectory radius \(r_m\) (both in meters) is:
\begin{equation}
     r_m = r_{gt} + \xi.
\end{equation}
 For prismatic joints, the relationship between the actual handle trajectory angle \(\theta_{gt}\) and the manipulated trajectory angle \(\theta_m\) (both in radians) is:
\begin{equation}
     \theta_m = \theta_{gt} + \xi.
\end{equation}
 For Bézier curves, the relationship between the actual control point position \(\boldsymbol{p}_{gt}\) and the manipulated trajectory control point position \(\boldsymbol{p}_m\) (both in meters), excluding the starting point, is:
\begin{equation}
     \boldsymbol{p}_m = \boldsymbol{p}_{gt} + 
    \begin{bmatrix}
        \xi &
        \xi
    \end{bmatrix}^{\transpose}.
\end{equation}

\paragraph*{ Impedance control}

 Impedance control, introduced by Hogan~\cite{hogan1985impedance}, is a robust technique widely used in robotic manipulation tasks. We compare impedance control with \method, following the same preliminary direction. In this context, impedance control serves as a prior-free approach for manipulating articulated objects and serves as a baseline in an ablation study to assess the essential role of tactile sensing in the \method.

\paragraph*{FlowBot3D}

 FlowBot3D, introduced by Eisner \etal~\cite{eisner2022flowbot3d}, is a recent advancement using modern learning techniques for generalized manipulation based on visual feedback. It employs a flow prediction model to predict part mobility and compute actions accordingly. It should be noticed that, to align with our simulation settings, we apply only the flow prediction model and action computation strategy. We assess its performance using the same tasks described in \cref{sec:sim_setup} as those used for \method, without additional fine-tuning.

\subsection{Results}\label{sec:sim-results}

\begin{figure*}[t!]
    \centering
    \includegraphics[width=0.98\linewidth]{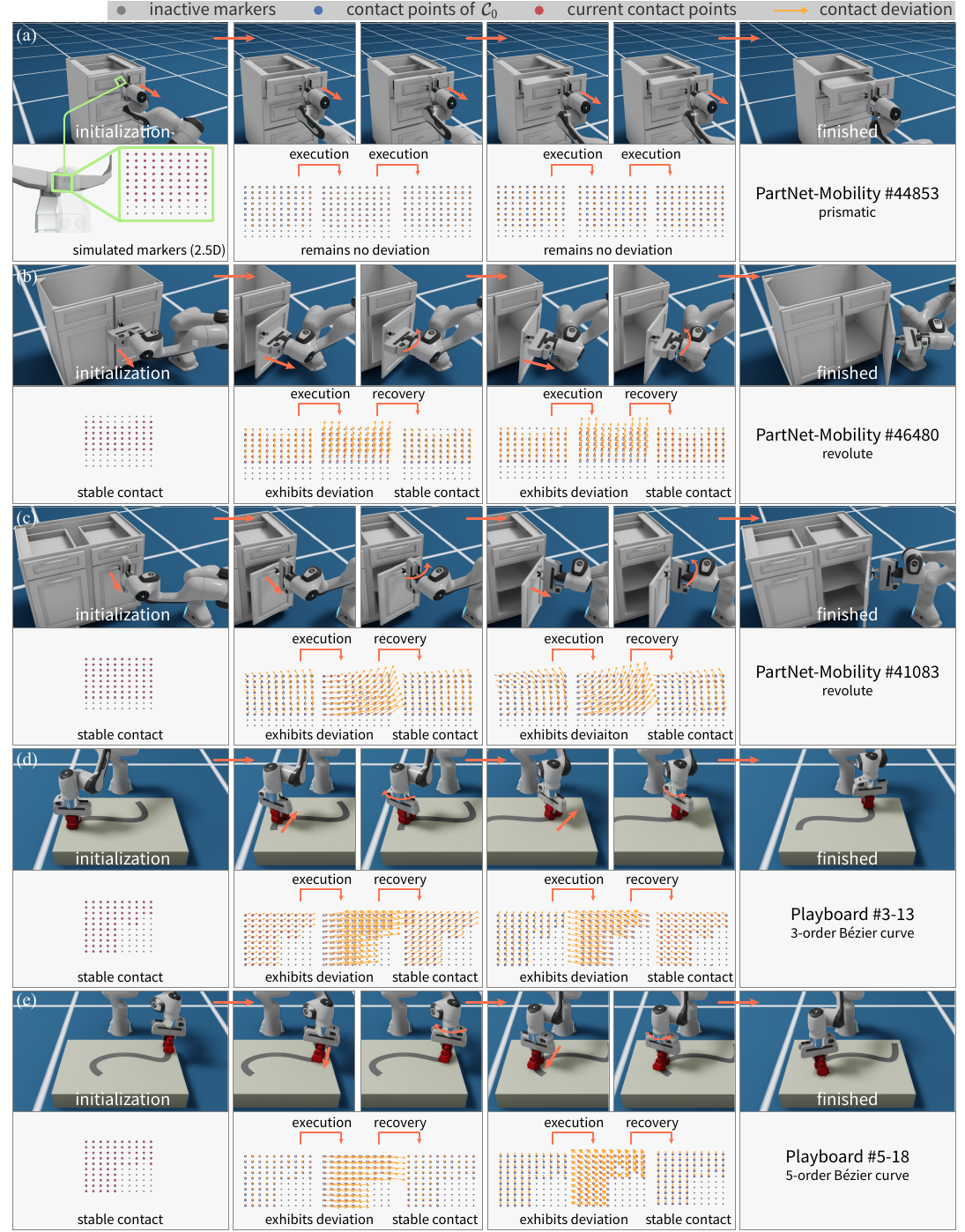}
    \caption{\textbf{Examples of simulated manipulation.} Each row captures a sequence of manipulating an articulated object, showcasing scenarios with basic joint types (a-c) and intricate trajectories (d-e). For each scenario, the depiction includes the initial and final states, the \textit{execution}-\textit{recovery} iterations, and the tactile patterns observed. This underscores the effectiveness of our simulation setup in demonstrating \method's capability to adapt and succeed across varying articulation challenges. For video demonstrations, refer to \href{\suppUrlSV}{Supp Video S5}.}
    \label{fig:sim-manipulation}
\end{figure*}

Our simulation results substantiate the efficacy of the setup in evaluating \method. In \cref{fig:sim-manipulation}, we highlight five representative examples: (a--c) illustrate the manipulation of objects with standard prismatic and revolute joints, while (d--e) detail scenarios involving intricate trajectory manipulations. The direction of pull and subsequent adjustments are indicated by the orange arrows. The simulated tactile feedback is visualized in the grey area beneath each figure, in which inactive, reference and current contact points are depicted as grey dots, blue circles, and red dots, respectively. Orange arrows, enlarged five-fold for clarity, depict the displacement from reference to current markers, with their size increasing in proportion to the error magnitude.

Consistent with our real-world experiments, deviations in the manipulation direction lead to deviation (\cref{fig:sim-manipulation}(a-c)) in the tactile pattern, signaling the need for corrective movements beyond the preliminary direction. Once the contact deviation surpasses \(\alpha \epsilon\) (\cref{fig:sim-manipulation}(b--e)), \method adeptly adjusts to re-establish stable contact.

\begin{figure}
    \centering
    \includegraphics[width=\linewidth]{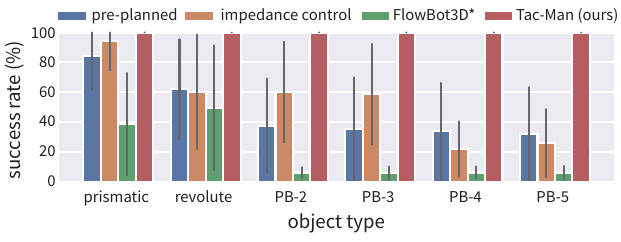}
    \caption{ \textbf{Barplot of the distance-weighted success rates of different methods across six articulated object types.} \method achieves \(100\%\) success rate, outperforming the other three baseline methods across all the articulated object types ($^*$ Tested on Isaac Sim using our settings without fine-tuning on PB-\(n\)).}
    \label{fig:sim-results}
\end{figure}

 The quantitative results for the three baseline methods and \method are summarized in \cref{fig:sim-results}. To provide a more informed measure of performance beyond binary success or failure, we introduce a distance-weighted success rate \(SR_W\), defined as:
\begin{equation}
     SR_W = \frac{\min(d_A, d_E)}{d_E} \times 100\%,
\end{equation}
 where \(d_E\) represents the expected contact point moving distance required for success, as specified in \cref{sec:sim_setup}, and \(d_A\) is the actual handle moving distance calculated from the final handle position in each trial.

 \cref{fig:sim-results} presents both mean and standard deviation of the weighted success rate across objects under different categories. Specifically, columns \(1\)-\(2\) display weighted success rates for manipulating objects with prismatic and revolute joints, while columns \(3\)-\(6\) show performance on intricate manipulation trajectories, with PB-\(n\) indicating performance on Bézier trajectories of order \(n\). Notably, \method achieves a \(100\%\) weighted success rate across all objects with various handle designs, markedly outperforming the three baseline methods and verifying its effectiveness. Key observations for the baseline methods include:
\begin{itemize}[leftmargin=*,noitemsep,nolistsep]
    \item  Pre-planned trajectory methods: In scenarios lacking tactile feedback, these methods are vulnerable to noise, affecting their performance.
    \item  Impedance control: Following the same preliminary direction as \method, yet without object kinematics priors and tactile feedback, this method proves effective for objects with prismatic joints. However, objects requiring movements beyond a straight trajectory may quickly surpass the compliance limits, resulting in failures.
    \item  FlowBot3D: This method demonstrates effectiveness with familiar objects featuring prismatic and revolute joints. However, it underperforms on unfamiliar, intricate trajectories.
\end{itemize}

\section{Discussion}\label{sec:discussion}

\begin{figure*}[t!]
    \centering
    \includegraphics[width=\linewidth]{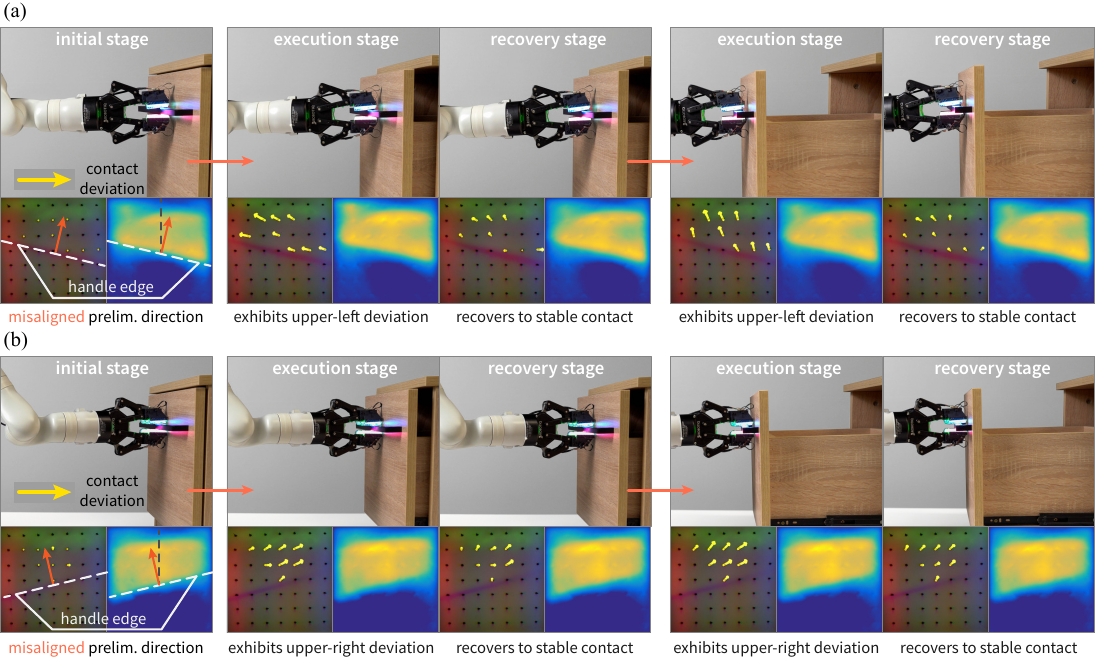}
    \caption{\textbf{Manipulation under imperfect initial grasp and misaligned preliminary direction.} Even given an imperfect initial grasp and a misaligned preliminary direction, \method is still able to use it to successfully open the drawer.}
    \label{fig:init_dir}
\end{figure*}

\subsection{Advancing Robotic Manipulation with \method}

Our study has validated the efficacy of \method, a tactile-informed, prior-free approach for manipulating articulated objects.  The core of \method is leveraging tactile feedback to discern discrepancies between the preliminary direction and actual interaction directions, informing dynamic adjustments to ensure precise manipulations. A standout feature of \method is its operational independence from pre-existing knowledge about object kinematics, presenting a considerable advantage over traditional methods reliant on priors. This capability is particularly beneficial in environments where priors may be ambiguous (\cref{sec:under-ambiguity}), imperfect (\cref{sec:under-uncertainties}), unknown (\cref{sec:intricate_traj}), or obsolescent (\cref{sec:under-perturbation})---conditions prevalent in dynamic, human-centric settings.  While classical control methods like impedance control can operate without object priors, they may fail with complex trajectories, as shown in \cref{fig:sim-results}.

\method enhances robotic autonomy by obviating the need for manually inputting accurate object kinematic models, a common requirement in earlier strategies~\cite{chitta2010planning,burget2013whole,jiao2021consolidating}. It sidesteps the limitations of methods that depend on visual perception for determining object kinematics~\cite{hu2017learning,abbatematteo2019learning,li2020category,zeng2021visual,mittal2022articulated,han2022scene,zhang2023part}, which are vulnerable to the ambiguities and inaccuracies intrinsic to visual data~\cite{zhu2020dark},  as demonstrated by the reduced effectiveness of pre-planned methods under noise in \cref{fig:sim-results}. Unlike approaches that utilize multi-frame observations to clarify ambiguities~\cite{hausman2015active,moses2020visual,martin2022coupled,eisner2022flowbot3d,lv2022sagci,jiang2022ditto,karayiannidis2012open,karayiannidis2016adaptive}, \method does not presuppose the existence of specific articulation types, thus demonstrating its versatility in managing novel articulation mechanisms as evidenced in \cref{sec:intricate_traj}  and \cref{fig:sim-results}.

Moreover, \method circumvents the exhaustive data collection process typically required by learning-based methods to tackle articulated object manipulation challenges~\cite{pastor2009learning,welschehold2017learning,zhang2018deep,lynch2020learning, xiong2021learning,qin2022dexmv,wong2022error,gong2023arnold,ye2023learning,urakami2019doorgym,xu2022universal,chen2022towards,geng2023partmanip,mo2019partnet,xiang2020sapien,liu2022akb,geng2023gapartnet}. The diversity and complexity of articulated objects, ranging from basic prismatic and revolute joints to more intricate mechanisms featuring simultaneous translation and rotation, pose a significant challenge for data-driven approaches,  as shown by FlowBot3D's lower success rates on unseen, intricate trajectories. Yet, \method adeptly addresses these challenges without relying on extensive datasets, thereby facilitating autonomous exploration and data collection for previously unseen objects. This autonomous data-gathering capability enhances the data collection process's versatility and substantially reduces human labor.

Additionally, \method showcases remarkable adaptability to dynamic changes within environments, as explored in \cref{sec:under-perturbation}. Its ability to adjust to unforeseen human interventions underscores a vital strength for robots operating in human-centric environments, where static assumptions about object kinematics can lead to manipulation failures. This adaptability is essential for ensuring seamless human-robot coexistence, enabling robots to effectively navigate the inherent unpredictability of real-world settings.

\subsection{Limitations and Future Work}\label{sec:limitation}

\paragraph*{Compatibility across different sensors}

\method's effectiveness primarily stems from its ability to compensate for contact deviation, which is modeled in a geometric perspective so that both direction and magnitude of deviation can be obtained, as detailed in \cref{sec:pose_calc}. This paper demonstrates the implementation of \method with GelSight-based sensor, as one economic representative for those sensors with the ability to measure contact geometry. We would like to clarify that the high-resolution property of GelSight is not a major consideration in this work, as indicated in the setup for experiments. Actually, all these sensors can be effectively adopted within the \method framework and should have a very similar formulation, with only minor variation in the pre-processing procedure where contact deviation geometry is extracted from raw sensor data. For example, the TacTip family~\cite{ward2018tactip} uses an array of pins to detect surface deformations, which can be mapped to contact points in a manner similar to our current implementation. Therefore, one important future work will be to investigate whether the extraction of contact deviation geometry with different sensors can be generalized to some degree to improve the succinctness of \method formulation.

Force/torque sensors, while primarily measuring forces and torques, can also be adapted for use within the \method framework, with an extra step of calculating geometric contact deviation, and thus the magnitude for compensation, from the force/torque measurements by Hooke's Law~\cite{gould1994introduction}:
\begin{equation}
    \differential\boldsymbol{f} = K \differential\boldsymbol{c}
\end{equation}
where \(\boldsymbol{c}\in\mathbb{R}^6\) represents the geometric contact deviation, and 
\(\boldsymbol{f}\in\mathbb{R}^6\) the force/torque measurement. \(K\) refers to the contact stiffness matrix, which can be constant or varying depending on the interaction surface, and often needs proper calibration before use. Therefore, another interesting future direction will be the investigation of contact stiffness calibration, which may further motivate the possibility of exploring force-based formulation for \method, where obtaining the magnitude of contact deviation from force becomes a major point.

Therefore, \method in principle has the ability to accommodate various types of sensors. Conducting comparative studies to evaluate the impact of different sensor types on \method's performance in different scenarios remains an important area for future work. These investigations could assist in optimizing sensor selection under specific tasks and potentially lead to a more generalized and succinct formulation of \method that can seamlessly incorporate data from diverse sensor types. 

\paragraph*{ Initial grasp generation and preliminary direction acquisition}

The effectiveness of \method relies on obtaining a stable initial contact \(\mathcal{C}_1\) and having a preliminary (rough) direction to manipulate an articulated object. In this study, we manually establish the initial (rough) grasp and preliminary direction for all experiments. Crucially, \method's success stems from its ability to adjust control input on the fly based on tactile feedback; hence, this initial grasp need not be precise and can sometimes even be intentionally wrong. To demonstrate \method's resilience, we conducted experiments on a drawer (see \cref{fig:init_dir}), providing the robot arm with intentionally incorrect grasps and directions. Despite these inaccuracies, \method adapts and successfully completes the task of opening the drawer.
 
Although \method does not require prior knowledge, such as object kinematics and learning policy, which was once deemed essential in the manipulation of articulated objects, its dependence on initial grasp and preliminary direction highlights a limitation but also suggests a potential area for future research. Improving the precision of learning-based initial grasps and directions~\cite{mo2021where2act,li2023gendexgrasp,li2024grasp}, or developing capabilities within \method to autonomously refine them, could enhance its effectiveness and broaden its application across various scenarios involving articulated objects. These improvements could potentially lead to more adaptable and autonomous robotic manipulation systems.

\begin{figure}[t!]
    \centering
    \includegraphics[width=\linewidth]{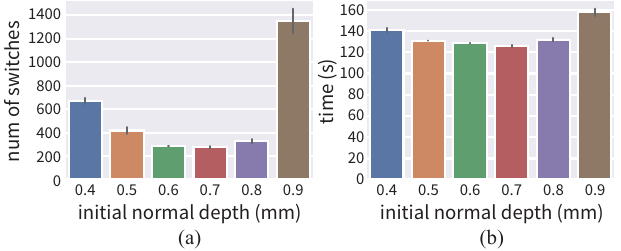}
    \caption{ \textbf{Impact of grasping force.} Extreme forces significantly reduce the efficiency of \method by affecting the computation of \(y\), the maximum distance during the execution stage, in the optimization program (7). This impact results in an increase in both the number of (a) execution-recovery cycles and (b) execution times. Crucially, within the elastic deformation region, \method yields successful manipulations of articulated objects despite varied execution time, demonstrating \method's robustness.}
    \label{fig:grasping-force}
\end{figure}

\paragraph*{ Grasping force}

 The initial grasping force critically influences \method's efficiency by directly impacting the stable contact state \(\mathcal{C}_1\) and associated tactile feedback. While it minimally affects the computation of the optimal transformation matrix in \cref{eq:stage2} (which relies on relative changes to \(\mathcal{C}_1\) for contact stabilization), it significantly influences the calculation of \(t\), the maximum distance during the execution stage. This \(t\) value, derived from the optimization program \eqref{eq:y}, depends on absolute changes. Simulation results in \cref{fig:grasping-force} demonstrate that both excessively large and small grasping forces prolong manipulation times. This aligns with intuition: larger forces may exceed the gripper's maximum elastic deformation capacity, while smaller forces increase slippage risk. Both scenarios are more likely to violate the stable contact constraint \eqref{eq:stable-contact}, reducing the maximum \(t\) value from program \eqref{eq:y} and necessitating additional iterations between execution and recovery stages. Crucially, within the elastic deformation region, \method yields successful manipulations of articulated objects despite varied execution time, demonstrating its robustness. Based on these findings, we practically select a grasping force that yields a deformation of approximately half the maximum normal elastic deformation \(e_n\). While this approach is effective, future work should explore methods for adaptively selecting the optimal grasping force to further enhance performance across various manipulation scenarios.

\paragraph*{Time efficiency}

\begin{figure}[t!]
    \centering
    \includegraphics[width=\linewidth]{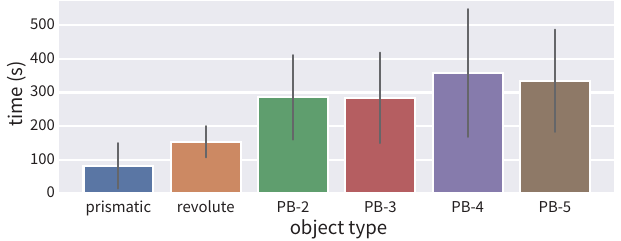}
    \caption{ \textbf{Time in second for completing tasks in simulation studies.}}
    \label{fig:sim-time}
\end{figure}

While the \method demonstrates notable capabilities in the prior-free manipulation of articulated objects, an area for potential enhancement is its time efficiency. As indicated in \cref{tab:time} and \cref{fig:sim-time}, the completion time for manipulation tasks using \method tends to exceed that of human performance. This disparity is largely attributed to the current implementation of \method, which operates without the benefits of closed-loop control.

\begin{table}[ht!]
    \centering
    \small
    \setlength{\tabcolsep}{2pt}
    \caption{\textbf{Time in second for completing tasks in the real world.}}
    \label{tab:time}
    \resizebox{\linewidth}{!}{%
        \begin{tabular}{cccccccc}
            \toprule
            \textbf{Task} & \cref{fig:under-ambiguity}(e) & \cref{fig:under-ambiguity}(f) & \cref{fig:under-ambiguity}(g) & \cref{fig:under-ambiguity}(h) & \cref{fig:uncertain_parameter}(b) & \cref{fig:intricate_traj}(a) &  \cref{fig:intricate_traj}(b)\\
            \midrule
            \textbf{Time (s)} & $\approx 480$  & $\approx 600$ & $\approx 480$ & $\approx 180$ & $\approx 750$ & $\approx 280$ &  $\approx 900$ \\
            \bottomrule
        \end{tabular}%
    }%
\end{table}

In its present form, \method employs a cautious, step-by-step exploratory approach. This method, devoid of pre-established object priors or learning, requires careful evaluation of, and response to, tactile feedback. The task duration figures reported in \cref{tab:time} and \cref{fig:sim-time} primarily reflect this initial phase of exploration, critical for the robot to accurately interpret and interact with the object.

Following this initial phase, the robot accumulates insights into the object's kinematics, aiding in the development of effective priors for future manipulations. These priors facilitate quicker subsequent manipulations, yet the absence of closed-loop control could lead to inefficiencies, particularly when initial conditions vary between attempts.

Looking ahead, enhancing \method to include velocity-based control elements could significantly improve its time efficiency. Such a development would enable quicker and more responsive adaptations during initial exploration, potentially reducing the overall time required for manipulations. It would lead to a more balanced approach, achieving manipulation robustness and efficiency on a par with human performance, especially in dynamic environments as discussed in \cref{sec:under-perturbation}.

\paragraph*{ Integrating \method into learnable framework}

 \method, as discussed in \cref{sec:tactile_servoing}, is a model-based tactile servoing method. This approach offers benefits such as reduced data collection requirements and enhanced explainability~\cite{edmonds2019tale}. However, integrating \method into a learnable framework represents a promising avenue for future research. This integration aims to generalize the strategy of utilizing tactile sensing in articulated object manipulation to a broader range of challenges, including dexterous manipulation~\cite{billard2019trends}, cable following~\cite{she2021cable}, surface tracking~\cite{lloyd2023pose}, and others.

\section{Conclusion}\label{sec:conclusion}

This paper introduces a novel prior-free approach to the manipulation of articulated objects, characterized by its reliance on tactile feedback rather than predefined object models. Central to this approach is the regulation of stable contact throughout the manipulation process, enabling dynamic adjustments of the robot's grip to ensure successful interaction with various objects.

The efficacy of \method is demonstrated through a series of experiments and simulations, highlighting its robustness in diverse scenarios. This includes its ability to handle ambiguous perceptions, adapt to imperfect parameter estimates, and respond effectively to unforeseen perturbations. The versatility of \method is further evidenced by its ability to manage objects with intricate trajectories in simulation.

In summary, this research makes a timely contribution to the field of robotic manipulation. It introduces a flexible and intuitive solution for interactions with articulated objects in environments marked by complexity and uncertainty. The adoption of a prior-free methodology marks a step towards robotic systems that are more adaptable and capable of performing sophisticated manipulation tasks, paving the way for enhanced automation and efficiency in various applications.

\section*{Acknowledgment}

We thank Hongjie Li (PKU), Yida Niu (PKU), and Zimo He (PKU) for setting up the experiments; Prof. Hao Dong (PKU), Zeyi Li (PKU), and Ruihai Wu (PKU) for setting up the Franka Robot; Dr. Chi Zhang (BIGAI) and Dr. Muzhi Han (UCLA) for insightful discussions; and Mr. Mish Toszeghi (QMUL) for meticulously proofreading our manuscript.

\appendix
\renewcommand\thefigure{A\arabic{figure}}
\setcounter{figure}{0}
\renewcommand\thetable{A\arabic{table}}
\setcounter{table}{0}
\renewcommand\theequation{A\arabic{equation}}
\setcounter{equation}{0}
\setcounter{footnote}{0}

\subsection{Details in Simulation Setup}\label{appendix:sim-details:setup}

\begin{table}[t!]
    \centering
    \small
    \setlength{\tabcolsep}{3pt}
    \caption{\textbf{Hyper-parameters defined in \cref{appendix:sim-details:setup,appendix:sim-details:gapartnet-preproc}}.}
    \label{tab:sim-hyper-params}
    \resizebox{\linewidth}{!}{%
        \begin{tabular}{ccccccccccc}
            \toprule
            \textbf{Para.} & \(n_\mathrm{res}\) & \(\prescript{\mathrm{sim}}{}\epsilon\) & \(\zeta_n\) & \(n_\mathrm{ctrl}\) & \(H\) & \(W\) & \(w\) & \(r_\mathrm{pin}\) & \(\eta\) \\
            \midrule
            \textbf{Value} & \(10\) & \(0.25~\mathrm{mm}\) & \(2~\mathrm{mm}\) & \(3\)-\(6\) & \(40~\mathrm{cm}\) & \(60~\mathrm{cm}\) & \(4~\mathrm{cm}\) & \(1.5~\mathrm{cm}\) & \(0.02\) \\
            \bottomrule
        \end{tabular}%
    }%
\end{table}

As detailed in \cref{sec:formulation}, \method exploits the compliance of the elastic deformable layer on the gripper to guide manipulation. This deformation provides critical tactile feedback by informing the gripper-object contact, which \method utilizes to adjust the gripper to maintain stable contact. Following previous studies~\cite{wang2022tacto}, we employ rigid-body simulations to efficiently and rigorously evaluate our method on a wide range of objects. Despite their limitation in handling deformable materials, we do not use soft-body materials due to much lower speed and higher computational demands. To closely replicate the tactile feedback mechanism under these constraints in the rigid-body simulation, we devised specific adaptations to estimate the behavior of a deformable layer, as detailed below. \(\prescript{\mathrm{sim}}{}(\cdot)\) denotes simulated variables.

\paragraph*{Simulating sensor markers}

We simulate the positions of the markers of the tactile sensors with point cloud grids. Specifically, we affix a point cloud grid, \(\prescript{\mathrm{sim}}{}{\mathcal{C}}_0 = \{ \boldsymbol{p}^0_i \vert i = 1, \dots, m \}\), on the grasping area of each gripper fingertip. We also add Gaussian noise \(\gamma \sim \mathcal{N} (0, \zeta^2)\) to the point cloud to imitate real-world sensor noise. 

\paragraph*{Simulating contact deviation}

Initially, the gripper stably grasps the handle, and the initial contact set \(\prescript{\mathrm{sim}}{}{\mathcal{C}_1}\) is determined as the markers close enough to the handle:
\begin{equation}
    \prescript{\mathrm{sim}}{}{\mathcal{C}_1} = \{ \boldsymbol{p}^{1} \vert d(\boldsymbol{p}^1) < \prescript{\mathrm{sim}}{}\epsilon\},
\end{equation}
where \(d(\boldsymbol{p})\) denotes the distance from point \(\boldsymbol{p}\) to the handle. This contact set is then transformed to the handle's frame \(\{h\}\) and affixed to it, denoted as \(\prescript{\mathrm{sim}}{}{\mathcal{C}_{h_1}}\). At each time step \(i\) during manipulation, the contact set \(\prescript{\mathrm{sim}}{}{\mathcal{C}_i}\) comprises the coordinated of the points affixed to the handle relative to the current gripper frame:
\begin{equation}
    \prescript{\mathrm{sim}}{}{\mathcal{C}_i} = \left\{\boldsymbol{T}_{h_i}^{i}\boldsymbol{p}^{h_i} + \begin{bmatrix}
    \gamma & \gamma & \gamma & 0
    \end{bmatrix}^{\transpose}\mid\boldsymbol{p}^{h_i}\in \prescript{\mathrm{sim}}{}{\mathcal{C}_{h_i}}\right\}.
\end{equation}
The discrepancy between \(\prescript{\mathrm{sim}}{}{\mathcal{C}_1}\) and \(\prescript{\mathrm{sim}}{}{\mathcal{C}_i}\) in the gripper's frame quantifies the contact deviation under the non-sliding stable contact conditions (\cref{eq:stable-contact}). Based on this information, the iterative process between the \textit{execution} and \textit{recovery} stages in the simulation adheres to the methodology detailed in \cref{sec:methods} and utilizes the same hyper-parameters as in the real-world experiments (\cref{tab:hyper-params}).

\paragraph*{Emulating the hypothetic deformable layer}

In rigid body simulations, we cannot simulate the elastic deformable layer, which causes misaligned dynamics compared to real-world scenarios. The fingers' movement directly applies frictional forces to the object handle to move it, rather than through the contact between the layer and the handle. This can lead to unrealistic movements of the object part during the recovery stage, where \cref{eq:formulation} does not guarantee a successful recovery. To address this discrepancy, we have empirically increased the damping within the joint settings during this stage to emulate the effect of the deformable layer. This modification suppresses the unintended object movement, aligning the simulated interactions more closely with those observed in real-world conditions. Although these adaptations do not perfectly mirror real-world trajectories, they provide a reasonable framework to assess how our algorithm tracks and maintains contact stability, thereby validating our approach.

\paragraph*{Determining hyper-parameters}

Key hyper-parameters play a significant role in our simulation studies, and their values are determined based on empirical observations and testing, which we report in \cref{tab:sim-hyper-params} to ensure transparency and reproducibility of the results.

\subsection{Preprocess PartNet-Mobility}\label{appendix:sim-details:gapartnet-preproc}

The PartNet-Mobility dataset~\cite{xiang2020sapien} comprises \(1045\) articulated objects, and GAPartNet~\cite{geng2023gapartnet} enriches it with detailed part-level annotations. In this dataset, each object includes a base link and one or more moving links, connected via prismatic or revolute joints. Objects with multiple joints often necessitate sequential manipulations to fully access all their links (\eg opening an oven door before pulling out the grill inside). For simplicity, we select one accessible handle on each of them for manipulation.

\subsection{Generate Random 1-\texorpdfstring{\ac{dof}}{} Playboards}\label{appendix:sim-details:playboard}

Our simulation setup incorporates playboards consisting of a baseboard (white) and a toy train (red), and the train is restricted to a predetermined trajectory (grey). To achieve this, we constructed a groove on a cuboid board and set up a distance joint between the board and the train, limiting its movement to within the plane of the board. Two cylindrical locating pins beneath the train's collision model are inserted into the groove to guide the train along the trajectory. The trajectories are sampled as Bézier curves, \(B(s), s \in [0, 1]\), with \(n_\mathrm{ctrl}\) control points uniformly sampled and scaled to fit an \(H \mathrm{cm} \times W \mathrm{cm}\) area, resulting in \((n_\mathrm{ctrl} - 1)\)-order curves. Any self-intersecting curves, taking into account a padding of \(w\), are discarded. The groove is carved into a \((H + 2w)~\mathrm{cm} \times (W + 2w)~\mathrm{cm}\) board following the curve. The initial position of the train is set at \(B(\eta)\) with a small positive \(\eta\) to prevent initial pin-border collisions. The hyper-parameters for this setup are listed in \cref{tab:sim-hyper-params}.

{
\bibliographystyle{ieeetr}
\bibliography{reference_header,reference}
}

\vskip -1\baselineskip plus -1fil
\begin{IEEEbiography}[{\includegraphics[width=1in,height=1.25in,clip,keepaspectratio]{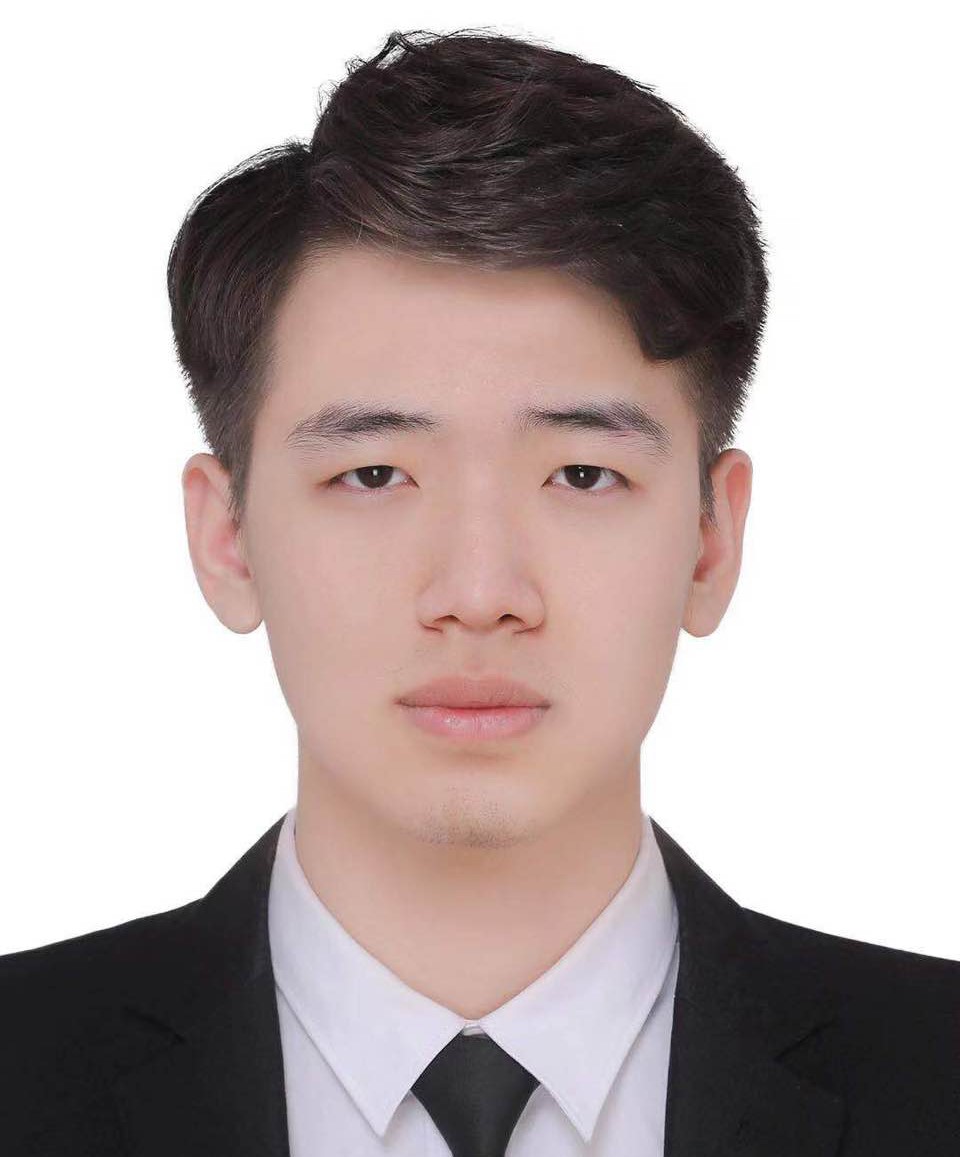}}]{Zihang Zhao}
received his B.Eng. degree from the School of Mechatronic Engineering at Harbin Institute of Technology in 2020, followed by an M.S. degree from the Department of Mechanical and Aerospace Engineering at the University of California, Los Angeles (UCLA) in 2021. He is currently pursuing his Ph.D. degree, advised by Prof. Yixin Zhu at the Institute for Artificial Intelligence, Peking University. His current research primarily focuses on interactive intelligence.
\end{IEEEbiography}

\vskip -3\baselineskip plus -1fil
\begin{IEEEbiography}[{\includegraphics[width=1in,height=1.25in,clip,keepaspectratio]{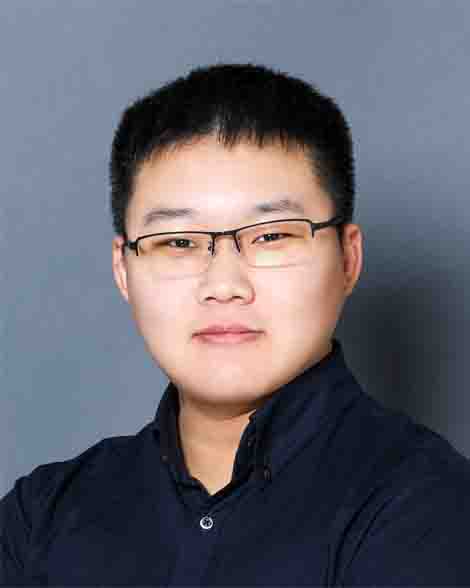}}]{Yuyang Li} is a first-year PhD student at the Institute for AI, Peking University. He is also a research intern at the Beijing Institute for General AI. He received his bachelor's degree in engineering from the Department of Automation at Tsinghua University in 2024. He aspires to advance the development of embodied intelligence, enabling machines to interpret multimodal sensory input, and interact with the environment with human-level adeptness.
\end{IEEEbiography}

\vskip -3\baselineskip plus -1fil
\begin{IEEEbiography}[{\includegraphics[width=1in,height=1.25in,clip,keepaspectratio]{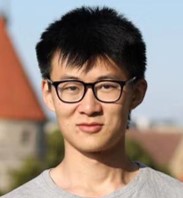}}]{Wanlin Li} received the B.Eng. and the Ph.D. degree from the School of Electronic Engineering and Computer Science at Queen Mary University of London (QMUL) in 2016 and 2021, respectively. He is currently a research scientist at the Beijing Institute for General Artificial Intelligence (BIGAI). His current research interests include force and tactile sensing.
\end{IEEEbiography}

\vskip -3\baselineskip plus -1fil
\begin{IEEEbiography}[{\includegraphics[width=1in,height=1.25in,clip,keepaspectratio]{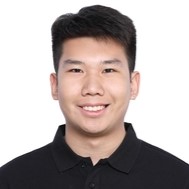}}]{Zhenghao Qi} is a sophomore student from the Department of Automation at Tsinghua University. He is also a student researcher in the Institute for AI at Peking University. His research interests include dexterous manipulation and robot motion control.
\end{IEEEbiography}

\vskip -3\baselineskip plus -1fil
\begin{IEEEbiography}[{\includegraphics[width=1in,height=1.25in,clip,keepaspectratio]{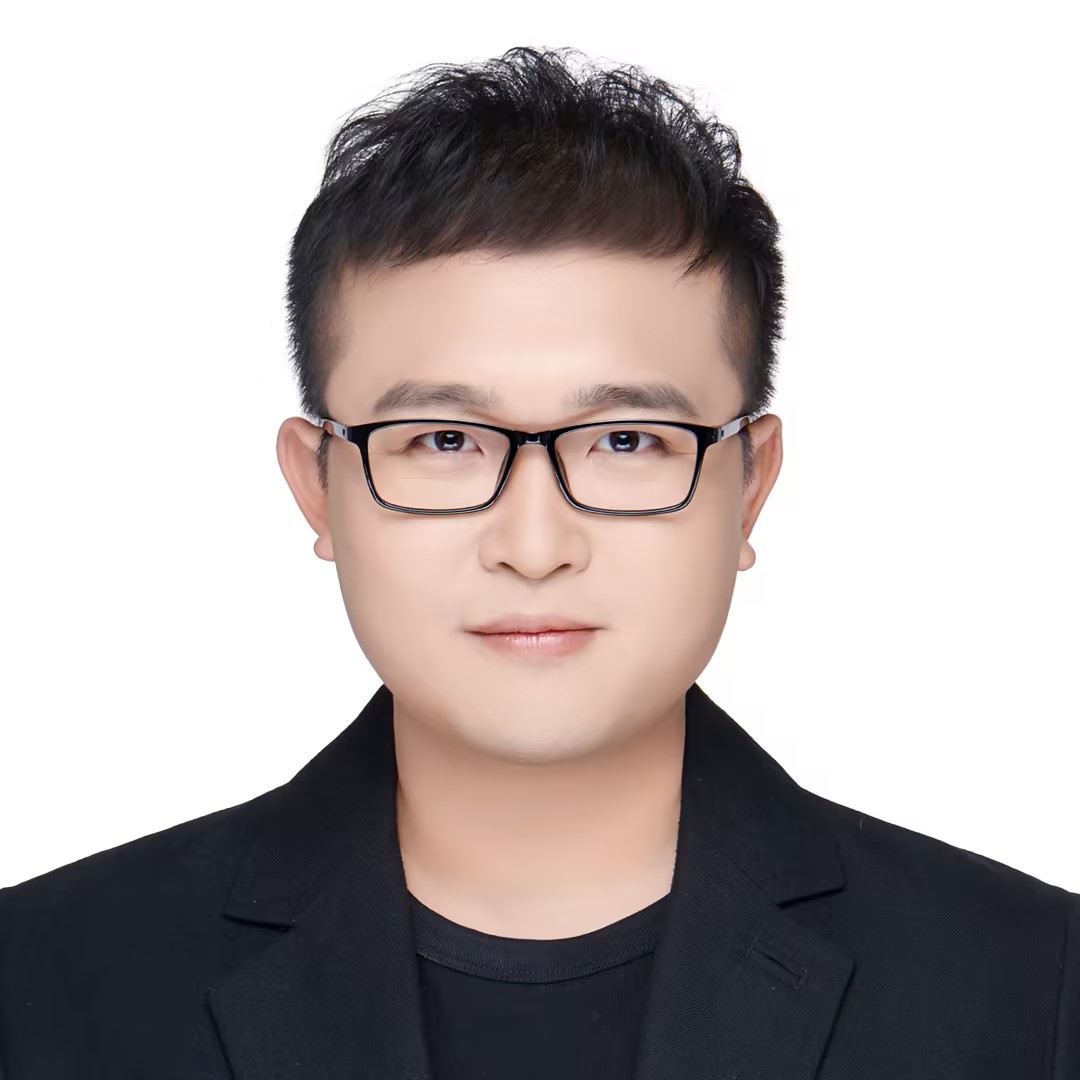}}]{Lecheng Ruan}
received the B.E. honor degree from the School of Mechatronic Engineering, Harbin Institute of Technology in 2015, and the Ph.D. degree from the Department of Mechanical and Aerospace Engineering, University of California, Los Angeles in 2020. He is now directing the Center of Intelligence in the Laboratory of Human-Robot Systems, Peking University. His research interests include perception, control and optimization of robotic systems.
\end{IEEEbiography}

\vskip -3\baselineskip plus -1fil
\begin{IEEEbiography}[{\includegraphics[width=1in,height=1.25in,clip,keepaspectratio]{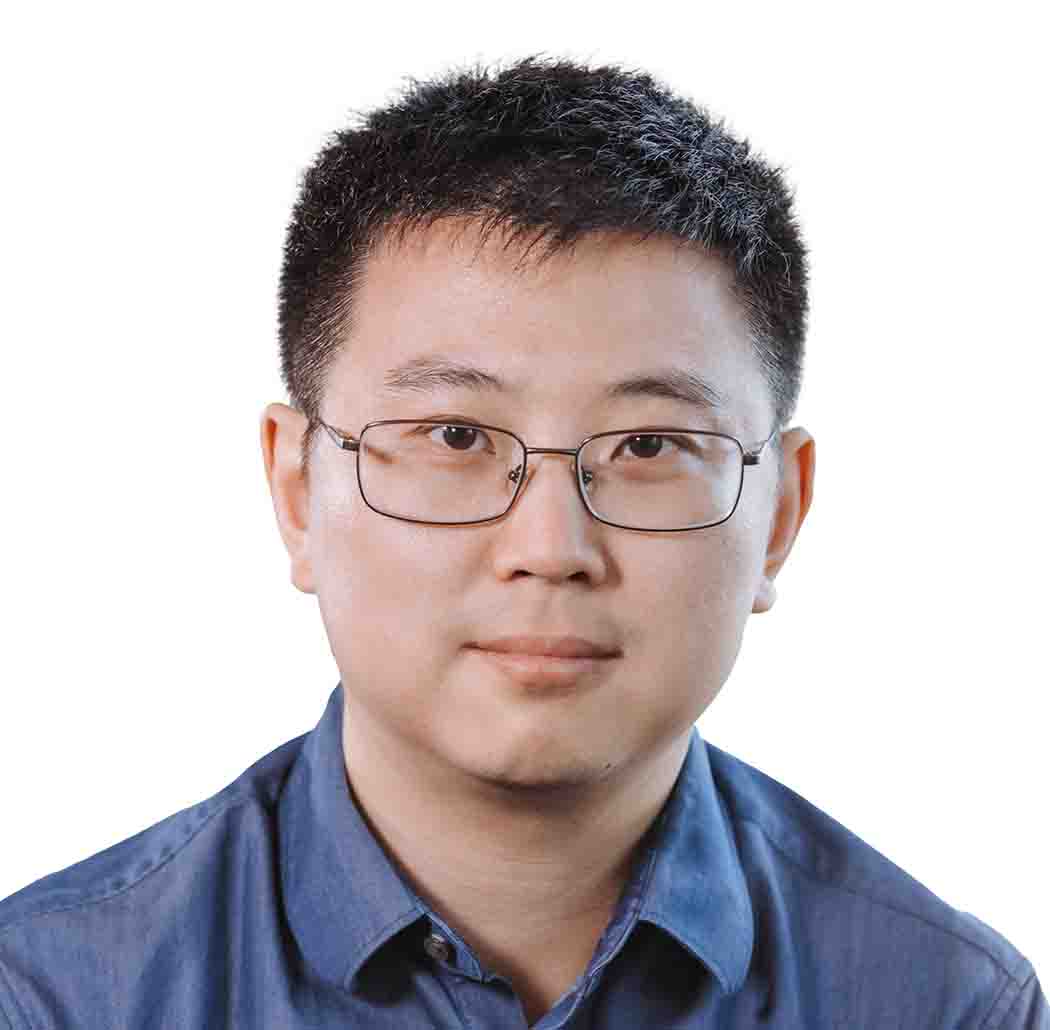}}]{Yixin Zhu}
is an Assistant Professor and the Assistant Dean at the Institute for Artificial Intelligence, Peking University. He received a Ph.D. degree from UCLA in 2018, advised by Prof. Song-Chun Zhu. His research builds interactive AI by integrating high-level common sense (functionality, affordance, physics, causality, intent) with raw sensory inputs (pixels and haptic signals) to enable richer representation and cognitive reasoning on objects, scenes, shapes, numbers, and agents.
\end{IEEEbiography}

\vskip -3\baselineskip plus -1fil
\begin{IEEEbiography}[{\includegraphics[width=1in,height=1.25in,clip,keepaspectratio]{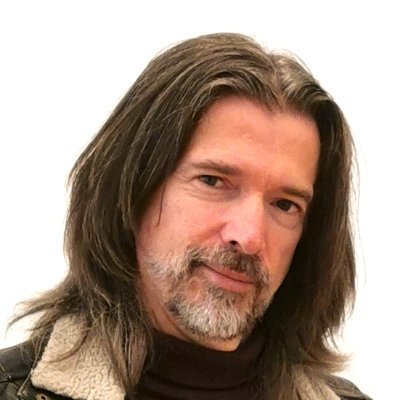}}]{Kaspar Althoefer}
(M'01-SM'19) is an expert roboticist conducting competitively funded research on soft robotics, intelligent tactile-sensing systems, and interaction dynamics modeling at Queen Mary University of London, with a focus on applications in minimally invasive surgery, assistive technologies, and manufacturing. Currently, he is a Professor of Robotics Engineering and Director of ARQ (Advanced Robotics @ Queen Mary) at Queen Mary University of London, U.K. He co-/authored more than 500 refereed research papers in mechatronics and robotics.
\end{IEEEbiography}

\end{document}